\documentclass[11pt,a4paper]{article}
\pdfoutput=1
\usepackage[hmargin=3.0cm,vmargin=3.0cm]{geometry}

\usepackage[table]{xcolor}  
\usepackage{graphicx}
\usepackage{caption,subcaption}
\usepackage{mathtools}
\usepackage{amsfonts}
\usepackage[inline]{enumitem}  
\usepackage{hyperref}
\usepackage[noabbrev]{cleveref}  
\hypersetup{
colorlinks=true,
citecolor=red
}

\usepackage{multicol}
\usepackage{tikz}
\usepackage{smartdiagram}
\usepackage[numbers]{natbib}  
\usepackage{authblk}            

\usepackage[space]{grffile}  
\graphicspath{{fig/}}  

\newcommand\zz{\mathbf{z}}

\newcommand\yy{\mathbf{y}}

\begin{document}

\title{Parametrization and generation of geological models with
  generative adversarial networks}

\author[]{Shing Chan\footnote{Corresponding author.\\ E-mail addresses: \texttt{sc41@hw.ac.uk} (Shing Chan), \texttt{a.elsheikh@hw.ac.uk} (Ahmed H. Elsheikh).} }
\author[]{Ahmed H. Elsheikh}
\affil[]{Heriot-Watt University, United Kingdom}
\affil[]{\small School of Energy, Geoscience, Infrastructure and Society}
\date{August 8, 2017\footnote{See also extended version (2019): \url{https://arxiv.org/abs/1904.03677}}}

\maketitle

\begin{abstract}

  One of the main challenges in the parametrization of geological
  models is the ability to capture complex geological structures often
  observed in the subsurface. In recent years, Generative
  Adversarial Networks (GAN) were proposed as an efficient method for
  the generation and parametrization of complex data, showing
  state-of-the-art performances in challenging computer vision tasks
  such as reproducing natural images (handwritten digits, human faces,
  etc.). In this work, we study the application of Wasserstein GAN for
  the parametrization of geological models.
  The effectiveness of the method is assessed for uncertainty
  propagation tasks using several test cases involving different
  permeability patterns and subsurface flow problems.
  Results show that GANs are able to generate samples that preserve
  the multipoint statistical features of the geological models both
  visually and quantitatively. The generated samples reproduce both
  the geological structures and the flow statistics of the reference
  geology.

\end{abstract}

\section{Introduction}
Generation and parametrization of geological models are fundamental
tasks in subsurface reservoir management. Due to inherent difficulties
in obtaining a complete and accurate description of the subsurface
properties -- such as the porosity and the permeability --,
engineers must contemplate a set of possible realizations of the
unknown and uncertain parameters while honoring the sparse information available
(e.g. information from well logs).
Standard algorithms used to generate geological realizations rely on the
variogram and two-point geostatistics,
however these algorithms are not suitable for generating complex
geological models that are often encountered in practice, such as
those containing channels and faults. To address this limitation, a number of
multipoint geostatistical simulators (MPS) were
introduced~\citep{strebelle2001reservoir,strebelle2002conditional,caers2004multiple}.
MPS methods infer properties from a conceptual model: Basically, a conceptual model is an
image designed under expert knowledge that conveys expected patterns of the
subsurface. Then MPS methods generate images that resemble the
conceptual model while honoring any other information available.

One limitation of MPS algorithms is that they do not provide
a useful parametrization of the output realizations. Concretely, for
efficient reservoir management it is of interest to be able
to characterize the geological realizations in terms of a few number
of parameters.
Moreover, it is often desirable that such parametrization be
differentiable. This is of practical importance because the generated
realizations are often coupled with other parts of the simulation
pipeline, e.g. for history matching or uncertainty quantification
tasks. In these procedures, the differentiability allows for efficient
inversion procedures as well as gradient-based optimization techniques.
%
%

The most standard approach in parametrization is principal component analysis
(PCA) (also called Karhunen-Loeve expansion), performed using a limited set
of realizations.
However, PCA is based on Gaussian assumptions and has limited capacity to
parametrize complex statistics such as those found in complex geological
structures with channelized features.
In the field of machine learning, where parametrization of high
dimensional data is ubiquitous, kernel principal component analysis
(kPCA)~\citep{scholkopf1998nonlinear} was introduced as an extension of PCA.
In essence, kPCA relies on an initial transformation of the input
data to enable more general features, followed by standard PCA
application on the transformed data. This is done indirectly through the
``kernel trick'' to avoid the evident computational implications of
transformations in very high dimensions.
The application of this machine learning tool in geostatistics was
first studied in~\citep{sarma2008kernel}, and further improved
in~\citep{ma2011kernel,vo2016regularized}.
Later, other PCA-based methods were proposed such as
in~\citep{vo2014new,vo2015data} where PCA is formulated as an
optimization problem, allowing the deliberate introduction of regularization
terms for conditioning and imposing spatial statistics. This was then
extended to kPCA methods in~\citep{vo2016regularized}.


In recent years, a novel parametrization method called \emph{generative
adversarial networks} (GAN)~\citep{goodfellow2014generative} was
introduced in the machine learning community. The formulation of GANs
takes a distinctive approach based on game theory, improving upon
previous generative models by avoiding intractable likelihood
functions and Markov chains. The sampling from the resulting generator
is performed simply using a forward function evaluation.

In this work, we study the application of Wasserstein
GAN~\citep{arjovsky2017wasserstein} (a particular implementation of GAN) for
the parametrization of complex geological models. The main motivation in the
study of GAN for the parametrization of complex geology comes from the well-known
capability of neural networks to learn very complex nonlinear features. 
In fact, in combination with convolutional
networks~\citep{lecun1989generalization,krizhevsky2012imagenet}, GAN models
have shown state-of-the-art performances in computer vision, specifically in
challenging tasks such as reproducing natural images (human faces,
handwritten digits, etc.~\citep{radford2015unsupervised}).
On a more related topic, GAN was recently employed in the
reconstruction of porous media~\citep{mosser2017reconstruction}. 
All this suggests that GAN can be effective in the parametrization
of very general geological models that exhibit complex structures.

The rest of this paper is organized as follows: In section 2, we
briefly describe parametrization of geological models using standard
principal component analysis. In section 3, we describe generative
adversarial networks and the Wasserstein formulation. In section 4, we
compare the generated realizations and present several numerical results in
subsurface flow problems. Finally, we draw our conclusions in section 5.

\section{Parametrization of geological models}
In a generic modeling setting, a spatial discretization grid is
defined over the physical domain of interest. The underlying random field
(e.g. porosity, permeability, etc.) then takes the form of a random vector
$\mathbf y = (y_1,\cdots,y_M)$ where $y_i$ is the value at grid block
$i$, and $M$ is the total number of grid blocks.
The goal of parametrization is to obtain a functional relationship
$\mathbf y=G(\mathbf z)$, where $\mathbf z = (z_1,z_2,\cdots,z_m)$ is
a random vector with assumed prior distribution $p_{\mathbf z}$, of
size much smaller than $\mathbf y$ (i.e. $m\ll M$). Additionally,
it is desirable that $G$ be differentiable. New realizations of the
random vector $\mathbf y$ can then be generated by sampling 
the random vector $\mathbf z$ from the prior $p_{\mathbf z}$.

The traditional approach in parameterizing $\mathbf y$ is to perform a
principal component analysis (PCA). Given a set of $N$ centered
realizations (i.e. zero mean)
$\mathbf y_1, \mathbf y_2, \cdots, \mathbf y_N$, let
$\mathbf Y = [\mathbf y_1, \mathbf y_2, \cdots, \mathbf y_N]$ be the
$M\times N$ matrix whose columns are the realization
vectors.  The covariance matrix of the random vector $\mathbf y$ can
be approximated by
\begin{equation} \label{eq:covariance_matrix}
\mathbf C = \frac{1}{N}\mathbf Y \mathbf Y^T
\end{equation}
PCA assumes the following parametrization of $\mathbf y$:
\begin{align}\label{eq:pca_decomposition}
\mathbf y(\boldsymbol \xi) &= \mathbf U \mathbf \Lambda^{1/2} \boldsymbol \xi\\
                           &= \xi_1\sqrt{\lambda_1}\mathbf u_1 + \cdots + \xi_M\sqrt{\lambda_M}\mathbf u_M \nonumber
\end{align}
where $\mathbf U = [\mathbf u_1,\cdots,\mathbf u_M]$ is the matrix
whose columns are the eigenvectors of the covariance matrix $\mathbf C$,
$\mathbf \Lambda$ is a diagonal matrix containing the respective
eigenvalues $\lambda_1,\cdots,\lambda_M$,
and $\boldsymbol \xi = (\xi_1,\cdots,\xi_M)$ is the parameterizing
random vector of uncorrelated components.
Assumming $\lambda_1 \geq \lambda_2 \geq \cdots \geq \lambda_M$, the
dimensionality reduction is obtained by keeping only the first $m$
components, that is, $\mathbf y(\mathbf z) = \xi_1\sqrt{\lambda_1}\mathbf u_1 + \cdots +
\xi_m\sqrt{\lambda_m}\mathbf u_m$ where
$\mathbf z = (\xi_1,\cdots,\xi_m)$, $m<M$.

It can be easily verified that realizations generated
by~\Cref{eq:pca_decomposition} reproduce the covariance matrix
$\mathbf C$. However, these new realizations do not in general
reproduce higher order moments of the original realizations.

\section{Generative adversarial networks}
Generative adversarial networks (GAN)~\citep{goodfellow2014generative}
consist of two competing functions (typically neural networks), namely
a generator $G$ and a discriminator $D$. 
The generator $G$ maps from input random \emph{noise} vector
$\mathbf z$ to output random vector $\mathbf y$ representing the
``synthetic'' images.
A prior distribution $p_{\mathbf z}$ is assumed for $\mathbf z$,
usually the standard normal distribution or an uniform distribution.
Given a dataset of original realizations $\mathbf y_1, \mathbf y_2, \cdots$
that are assumed to come from an unknown distribution $p_{\textrm{data}}$,
the goal for $G$ is to sample from this distribution -- informally, to
generate samples that resemble samples from the dataset.
Given that $p_\zz$ is prefixed, $G(\zz)$ induces a distribution $p_G$ that
depends only on (the weights of) $G$.
The objective is to optimize $G$ so that $p_G=p_{\textrm{data}}$.
On the other hand, the discriminator $D$ is a classifier function
whose goal is to determine whether a realization $\yy_i$
is real (i.e. coming from the dataset) or synthetic (i.e. coming
from the generator). The output of $D(\mathbf y_i)$ is a
scalar representing the probability of $\mathbf y_i$ coming from the
dataset.

In short, the goal of $D$ is to correctly discern between synthetic
and real samples, while the goal of $G$ is to fool $D$. This
interplay between $G$ and $D$ is formulated as a two-player
minmax game:
\begin{equation}
  \label{eq:minmaxgame}
  \min_G \max_D \{ \mathbb E_{\mathbf{y}\sim p_{\mathrm{data}}}{\log D(\mathbf y)} + \mathbb E_{\mathbf{z}\sim p_{\mathbf z}}{\log(1-D(G(\mathbf{z})))} \}
\end{equation}

In practice, the solution to this game can be approached by
iteratively optimizing $G$ and $D$ in an alternating manner and
independently. Given infinite capacity, it is shown in~\citep{goodfellow2014generative} that the game minimizes the Jensen-Shannon divergence between $p_G$ and $p_{\mathrm{data}}$, with global
optimum $p_G=p_{\mathrm{data}}$ when $D=\frac{1}{2}$
everywhere in its support. That is, the generator $G$ learns to replicate the data
generating process, therefore the discriminator
$D$ fails to distinguish real samples from synthetic ones, resulting in a
``coin toss'' situation.
Once trained $G$ provides the achieved parametrization and $\mathbf z$ the
reduced representation. We note that the differentiability of $G$
with respect to $\mathbf z$ is a direct consequence of choosing a family of
differentiable functions for $G$, which is normally the case when neural networks
are employed.

\subsection{Wasserstein GAN}
Finding the equilibrium of the game defined by~\eqref{eq:minmaxgame} is
notoriously hard in practice. In particular, if the discriminator $D$ is
``too well trained'', it becomes saturated and provides no useful information
for improvement of $G$. Another issue is mode collapse, where the generator
generates only a single image that $D$ always accepts. Solving these issues
has been the focus of many recent works, as summarized
in~\citep{salimans2016improved,arjovsky2017towards}.

One major development in improving GANs is Wasserstein GAN
(WGAN)~\cite{arjovsky2017wasserstein}. In WGAN, the focus is in
directly minimizing the Wasserstein distance instead of the
Jensen-Shannon divergence. The Wasserstein distance is defined as:
\begin{equation}
  \label{eq:wasserstein}
  W( p_{\textrm{data}}, p_G) =
  \inf_{\gamma\in\Pi(p_{\textrm{data}}, p_G)}{
    \mathbb E_{(\mathbf y_1,\mathbf y_2)\sim\gamma}{||\mathbf y_1-\mathbf y_2||}
  }
\end{equation}
where $\Pi(p_{\textrm{data}}, p_G)$ denotes the set of all joint distributions
$\gamma(\mathbf y_1,\mathbf y_2)$ with marginals $p_{\textrm{data}}$ and $p_G$,
correspondingly.
The Wasserstein distance as expressed
in~\Cref{eq:wasserstein} is intractable in practice, therefore the
following expression, due to the Kantorovich-Rubinstein duality, is
used instead:
\begin{align}
  \label{eq:wasserstein_sup}
  W(p_{\mathrm{data}}, p_G) &= \frac{1}{K}\sup_{||f||_L\leq K}\{\mathbb E_{\mathbf y\sim p_{\mathrm{data}}}{f(\mathbf y)} - \mathbb E_{\mathbf y\sim p_G}{f(\mathbf y)} \} \nonumber\\
                                   &= \frac{1}{K}\sup_{||f||_L\leq K}\{\mathbb E_{\mathbf y\sim p_{\mathrm{data}}}{f(\mathbf y)} - \mathbb E_{\mathbf z\sim p_{\mathbf z}}{f(G(\mathbf z))} \}
\end{align}
where $||f||_L\leq K$ indicates that the supremum is over the set of
real-valued $K$-Lipschitz functions, for some (irrelevant) constant
$K$.
In effect, the function $f$ in~\Cref{eq:wasserstein_sup}, also called the ``critic'', replaces the discriminator
$D$ in standard GAN. Instead of a classifier function $D$
outputing a probability, the critic $f$ outputs a general real-valued
score to the samples.

Even though the theoretical derivation of WGAN is very different from
standard GAN, the only difference in the resulting formulation is a
small change in the objective function, plus the Lipschitz constraint
in the search space of the adversarial function (the critic), although
relaxing the requirement to be a classifier function.
However, the practical implications of WGAN are significant:
it is much more stable and it provides a
meaningful loss metric during training that correlates with the quality of
the generated samples. This is very important for assessing the training
progress, as well as providing a criteria for convergence.
For more detailed discussion, see~\citep{arjovsky2017wasserstein}.

\subsubsection{Training WGAN}
Typically, both the critic function $f$ and the generator $G$ are
neural networks with specified architectures. Let $ w$ and
$ \theta$ represent the weights of the networks for $f$ and
$G$, respectively, and we use the symbols $f_{ w}$ and
$G_{\theta}$ to explicitly express this dependency. To
enforce the Lipschitz condition of $f_{ w}$, it is enough to
assume a compact search space $\mathcal W$ for the weights
$ w$. In practice, this can be done by clipping
the values of $ w$ within certain range after each update of the
training iteration.

The training procedure consists of alternating steps where $G$ and $f$
are optimized independently: In step \texttt{A}, we fix the generator weights
$\theta$ and optimize the critic network
$f_{ w},\;  w \in \mathcal W$ (ideally to optimality)
following the maximization problem:
\begin{equation}\label{eq:wasserstein_max}
  \max_{ w \in \mathcal W}\{
  \mathbb E_{\mathbf y\sim p_{\mathrm{data}}}{f_{ w}(\mathbf y)} - \mathbb E_{\mathbf z\sim p_{\mathbf z}}{f_{ w}(G_{\theta}(\mathbf z))}
  \}
\end{equation}
Once the critic is fitted, we are provided with an approximation of
the Wasserstein distance $W(p_{\mathrm{data}}, p_{G_{\theta}})$ (up
to a multiplicative constant).
In step \texttt{B}, this
distance is differentiated with respect to $\theta$ to
obtain the gradient required to optimize $G_{\theta}$:
\begin{equation} \nabla_{\theta} W(p_{\mathrm{data}},
p_{G_{\theta}}) = -\mathbb E_{\mathbf z\sim p_{\mathbf
z}}{\nabla_{\theta}f_{ w}(G_{\theta}(\mathbf z))}
\end{equation}
Steps \texttt{A} and \texttt{B} are iteratively repeated until
convergence is reached.

The quality of the approximation of the Wasserstein distance, and
therefore of its gradient, will depend of course on how well the
critic has been trained. Hence, it is desirable to train the critic
relatively well in each iteration. In practice, it is common to fully
optimize $f$ every so often by setting a high number of inner
iterations in step \texttt{A} after a number of full
\texttt{A}-\texttt{B} loops. This is in contrast to standard GAN where
the training of the discriminator $D$ and the generator $G$ has to be
carefully balanced. Moreover, the readily available approximated
Wasserstein distance provides an useful metric that correlates well
with the quality of generated samples. We note the lack of a similar a
metric in the standard GAN formulation.

\section{Numerical Experiments}

In this section, we demonstrate the use of Wasserstein GAN (
referred simply as GAN in the rest of this section) for the parametrization
of two permeability models. We assess the generated
realizations in subsurface flow problems.  An uncertainty
propagation study is performed to estimate densities for several
quantities of interest.  Results are compared with the true results
derived from the original realizations. We also include results based on
standard PCA parametrization.

\subsection{The dataset}
\begin{figure} \centering{
  \begin{subfigure}{.4\linewidth} \centering
    \includegraphics[width=\linewidth]{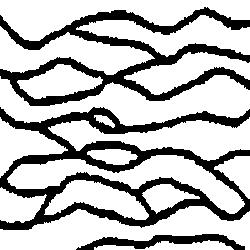}
    \caption{\emph{Semi-straight} channels}
    \label{fig:channel}
  \end{subfigure}
  \begin{subfigure}{.4\linewidth} \centering
    \includegraphics[width=\linewidth]{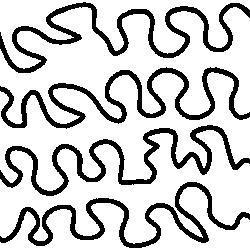}
    \caption{\emph{Meandering} channels}
    \label{fig:meandering}
  \end{subfigure}
  \begin{subfigure}{\linewidth} \centering
    \includegraphics[width=.50\linewidth]{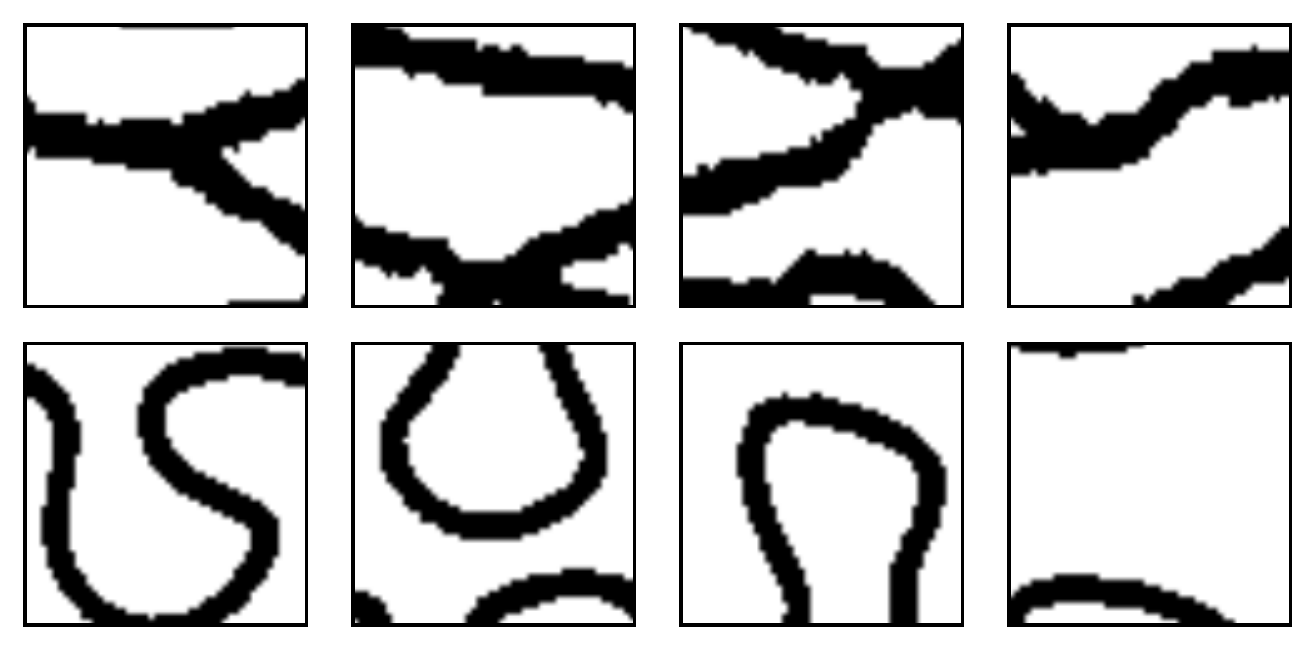}
    \caption{Samples obtained by cropping}
    \label{fig:inputs}
  \end{subfigure}
}
  \caption{Conceptual images (a) and (b) employed to generate the set
    of realizations (c). The size of the conceptual images are
    $250\times 250$, and the size of the realizations are
    $50\times 50$.}
\end{figure}

We evaluate the effectiveness of GANs in parameterizing two
permeability models. The first model is characterized by
channelized structures showing \emph{semi-straight} patterns. The
second model consists of channelized structures exhibiting heavy
\emph{meandering} patterns.
Conceptual images\footnote{Conceptual images or conceptual models are
  also called \emph{training images} in the geology
  literature. Unluckily, the same term is also employed in machine
  learning to refer to the input training images (the
  realizations). Hence, we do not employ such term to avoid
  confusion.}  for each permeability type are shown in
\Cref{fig:channel,fig:meandering}. These are binary images
representing log-permeability values, where $1$ (black) designates
the high permeability channels and $0$ (white) designates the low
permeability background. In practice, these could be channels of sand
embedded in a background of clay. The conceptual images used are of
size $250\times 250$.

We consider realizations of the permeability of size $50\times
50$. Typically, these are obtained by passing the large conceptual image
to a multipoint geostatistical simulator such as
\texttt{snesim}~\citep{strebelle2001reservoir} which will generate a dataset of
realizations.
Instead, here we obtained our dataset by simply cropping out patches of size
$50\times 50$ from the conceptual images. A dataset of $40,000$ ``realizations'' can
be quickly obtained in this way from each model, although
with significant overlap between samples. Some samples are shown
in~\Cref{fig:inputs}.

\subsection{The setup}

\begin{figure}
  \centering{
  \begin{subfigure}{.95\linewidth}
    \includegraphics[width=\linewidth]{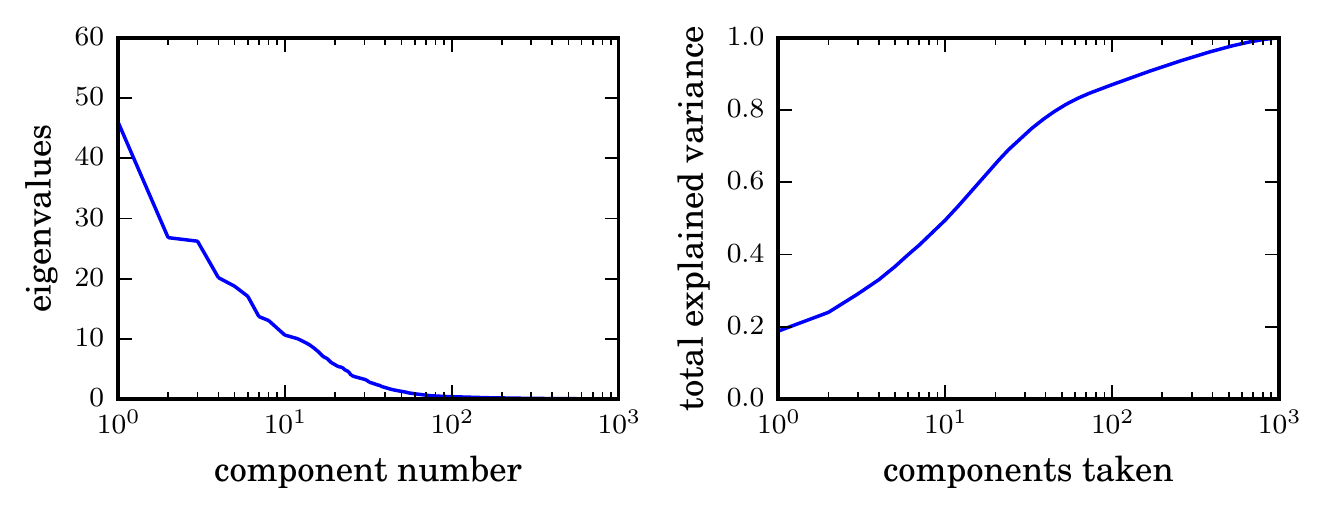}
    \caption{Eigenvalues and total variance explained in the
      \emph{Semi-straight} pattern}
  \end{subfigure}
  \begin{subfigure}{.95\linewidth}
    \includegraphics[width=\linewidth]{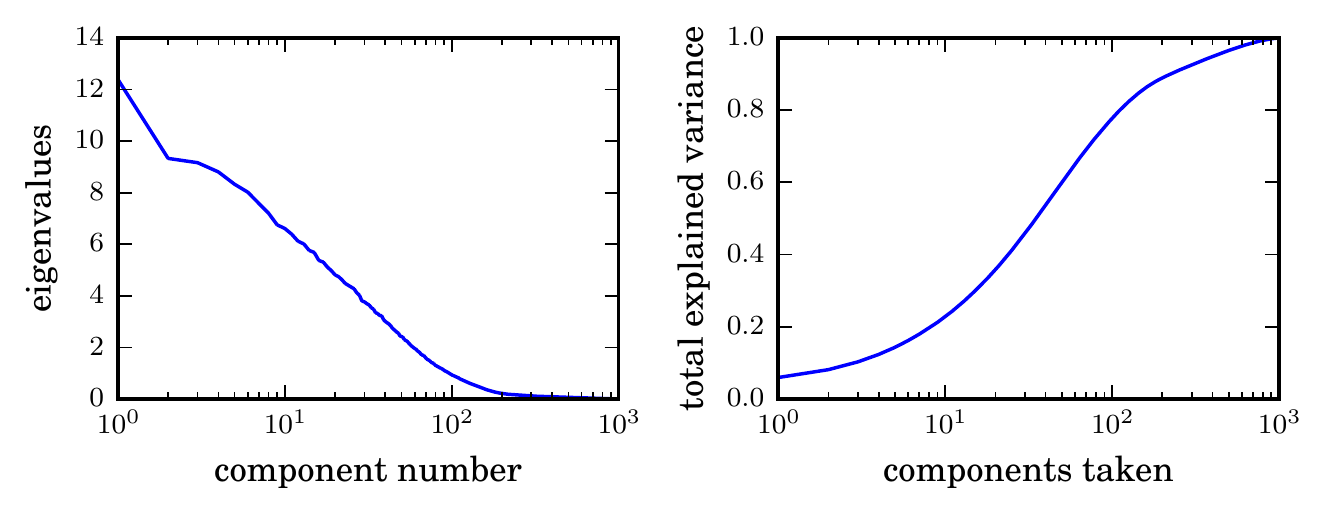}
    \caption{Eigenvalues and total variance explained in the
      \emph{meandering} pattern}
  \end{subfigure}
}
  \caption{Results from the principal component analysis. We retained
    $75\%$ of the total variance, which corresponded to retaining $37$
    components in the \emph{semi-straight} pattern, and $104$
    components in the \emph{meandering} pattern.}
  \label{fig:scree}
\end{figure}

For each pattern considered, a dataset of $40,000$ realizations is used
to train a GAN model and to perform PCA for comparison. For the design
of our GAN architecture, we followed the guidelines provided
in~\citep{radford2015unsupervised}. We found that the inclusion of
fully connected layers worked best in our experiments. More
specifically, we included one fully connected layer after the input of
$G$, and one before the output of $D$. The rest of the architecture design
follows the suggestions from the guidelines.
The network is trained with the Wasserstein formulation using the
default settings provided in~\cite{arjovsky2017wasserstein}.
For the size of the noise vector $\mathbf z$, we looked at sizes $20$
and $40$, and we assumed standard normal distribution for the
prior $p_{\mathbf z}$.
We used $\tanh$ as the output activation of the generator
$G$ and we preprocessed the binary images before training by
shifting and scaling, from the binary range $[0,1]$ to the
$\tanh$ range $[-1,1]$. Using $\tanh$ as the output activation
automatically bounds the output of the generator.
After the generator is trained and new realizations are
generated, we can simply transform the outputs back to the binary
range\footnote{Alternatively, one can use a sigmoid function, which
  is already bounded in $[0,1]$, as the output activation of
  $G$. However, $\tanh$ offers better properties during
  training, such as already providing zero-centered inputs for the
  discriminator.}.

Regarding PCA, we followed the procedure in published
literature~\citep{sarma2008kernel,ma2011kernel} of retaining $75\%$ of
the total variance. This corresponded to retaining $37$
eigencomponents in the semi-straight pattern, and $104$
eigencomponents in the meandering pattern. A scree-plot of the
eigenvalues is shown in~\Cref{fig:scree} for both permeability patterns.

\subsection{Comparing realizations}

\begin{figure} \centering{
  \begin{subfigure}{.95\linewidth}
  \includegraphics[width=\linewidth]{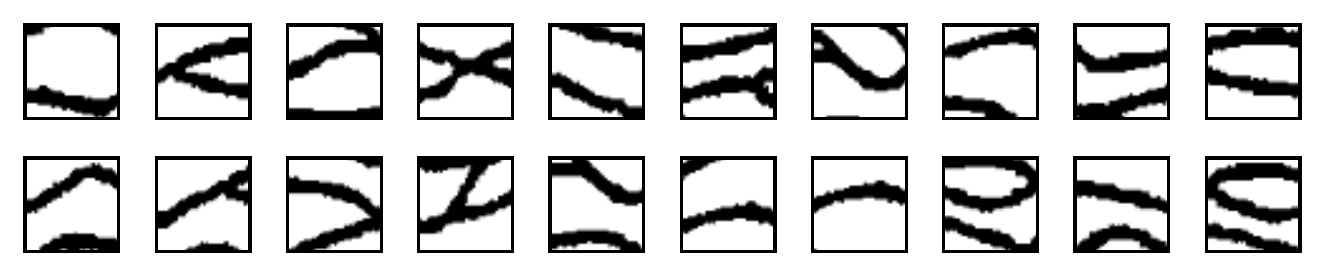}
  \caption{Original realizations}
  \end{subfigure}
  \begin{subfigure}{.95\linewidth}
  \includegraphics[width=\linewidth]{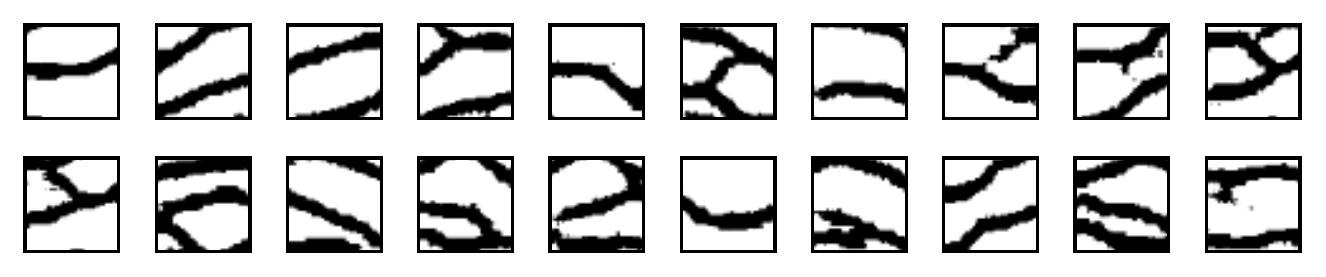}
  \caption{Realizations generated using GAN20}
  \end{subfigure}
  \begin{subfigure}{.95\linewidth}
  \includegraphics[width=\linewidth]{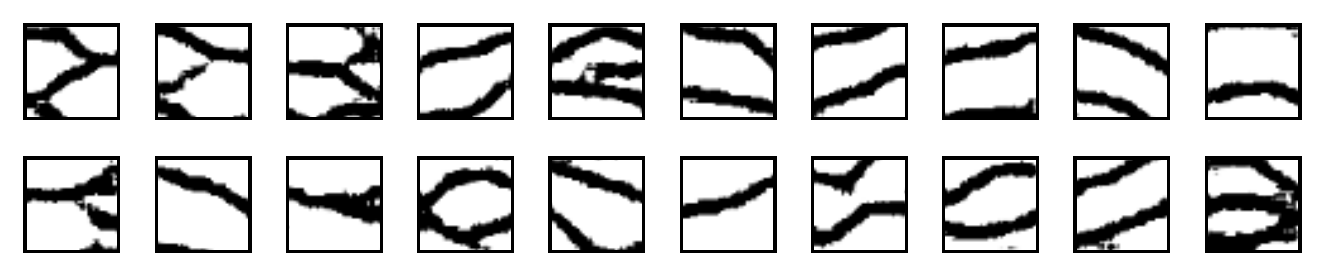}
  \caption{Realizations generated using GAN40}
  \end{subfigure}
  \begin{subfigure}{.95\linewidth}
  \includegraphics[width=\linewidth]{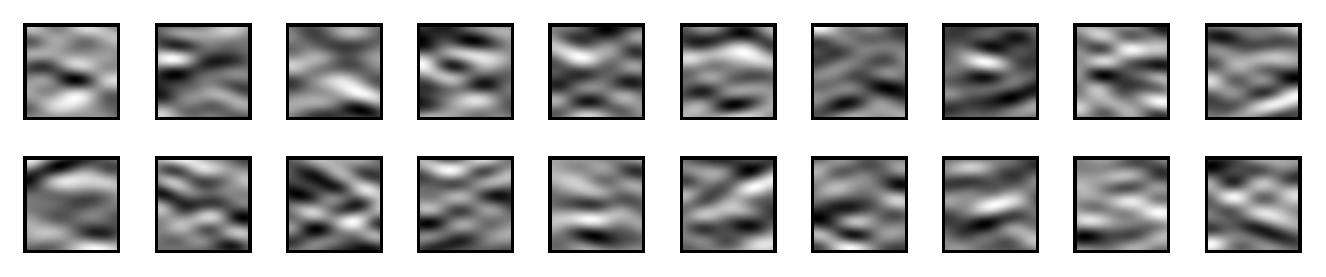}
  \caption{Realizations generated using PCA}
  \end{subfigure}
}
  \caption{Realizations for the \emph{semi-straight} channelized
    structure.}
  \label{fig:channel_outputs}
\end{figure}

\begin{figure} \centering{
  \begin{subfigure}{.95\linewidth}
  \includegraphics[width=\linewidth]{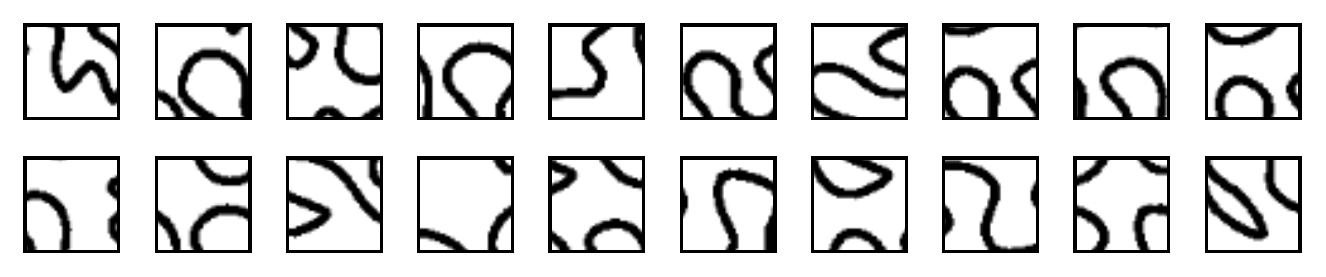}
  \caption{Original realizations}
  \end{subfigure}
  \begin{subfigure}{.95\linewidth}
  \includegraphics[width=\linewidth]{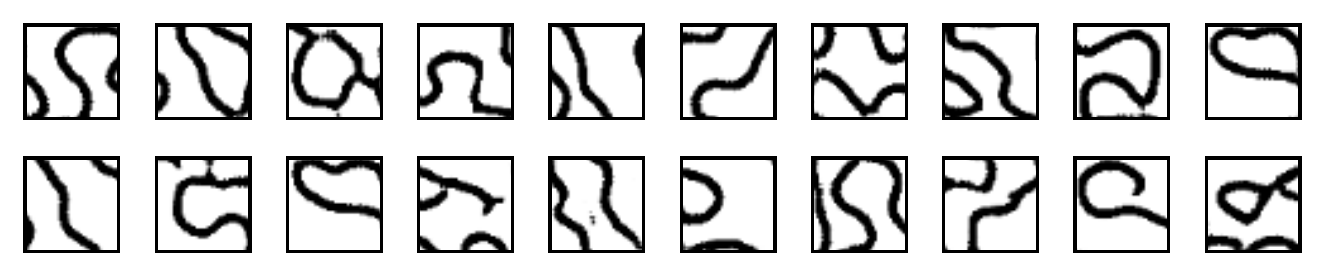}
  \caption{Realizations generated using GAN20}
  \end{subfigure}
  \begin{subfigure}{.95\linewidth}
  \includegraphics[width=\linewidth]{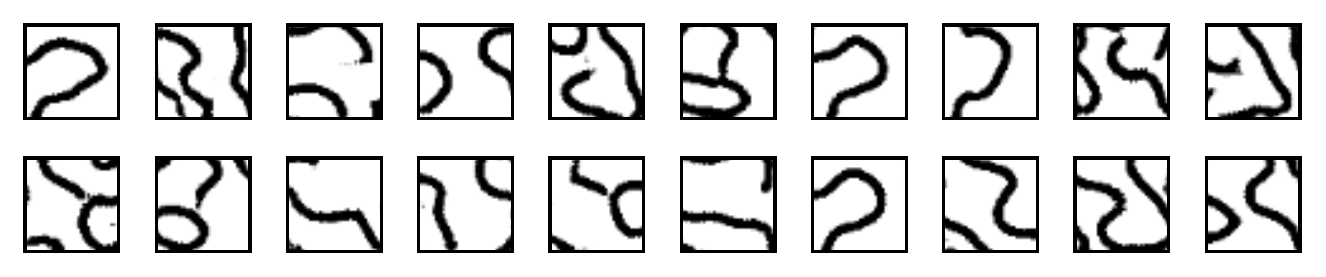}
  \caption{Realizations generated using GAN40}
  \end{subfigure}
  \begin{subfigure}{.95\linewidth}
  \includegraphics[width=\linewidth]{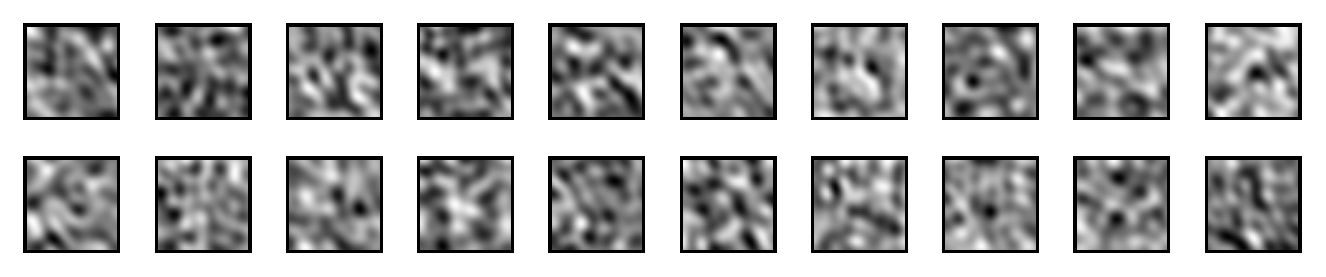}
  \caption{Realizations generated using PCA}
  \end{subfigure}
}
  \caption{Realizations for the \emph{meandering} channelized
    structure.}
  \label{fig:meandering_outputs}
\end{figure}

\begin{figure}
\centering{
  \begin{subfigure}{.95\linewidth}
    \includegraphics[width=\linewidth]{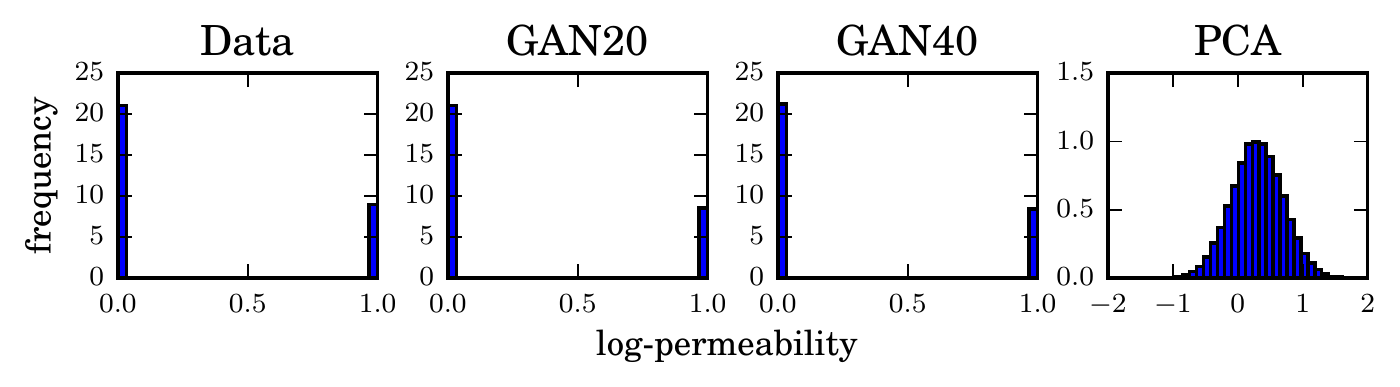}
    \caption{\emph{Semi-straight} pattern}
  \end{subfigure}
  \begin{subfigure}{.95\linewidth}
    \includegraphics[width=\linewidth]{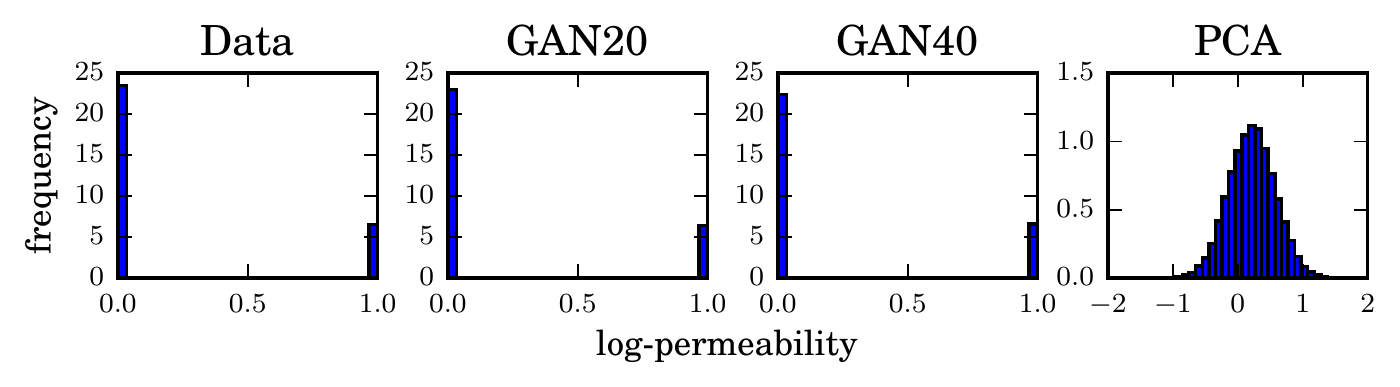}
    \caption{\emph{Meandering} pattern}
  \end{subfigure}
}
  \caption{Histogram of permeability values at the center of the
    domain.}
  \label{fig:perm_hist}
\end{figure}

We denote by GAN20 and GAN40 the GAN models trained
with input $\mathbf z$ of sizes $20$ and $40$, respectively.
\Cref{fig:channel_outputs,fig:meandering_outputs} show realizations
generated with the GAN models and PCA, along with original
realizations. These realizations are fair draws (not cherry-picked).
It is evident from these images that the GAN models outperform PCA in
capturing the channelized patterns of the data. In the semi-straight
case, PCA slightly manages to capture the horizontal correlation of
the channels, but it is clear that the generated samples are not
channelized. In the meandering case, PCA seems to completely fail in
capturing any pattern of the data.
Additional samples are displayed in~\Cref{sec:appendix}.

Visually, we did not see much difference in samples generated
by GAN20 and GAN40.  Overall, both models
generated realizations that are highly plausible, with samples that
may trick the human eye. There are, however, some features that expose
them such as isolated pixels, although this could be removed by applying a
median filter. Perhaps of more importance in regards to reservoir
simulation are the ``hanging channels'', i.e. channels with endpoints
inside the domain, which are not present in the original realizations.
In principle, this could be addressed by further tuning the
network architecture.
%

Following the visual comparison, we plot the histogram of the
permeability values at the center of the domain. For this, we generated a
number of realizations using the GAN models and PCA for each of the
permeability patterns. The results presented in~\Cref{fig:perm_hist} show
that both GAN20 and GAN40 learn the correct type of distribution (binary)
with the vast majority of the values toward the extremes ($0$ and $1$), and
the obtained frequencies are very close to data (original realizations). In
contrast, the results from PCA follow a normal distribution, as expected.

\subsection{Uncertainty propagation study}
In this subsection, we study the effectiveness of the parametrization
for an uncertainty propagation task.  Since no significant differences
were found in the visual comparison between GAN20 and GAN40, we only
consider the results for GAN20 to simplify the presentation. We denote
the results of GAN20 simply as GAN in the rest of this subsection.

We consider a subsurface flow problem where water is injected in an 
oil-filled reservoir. The physical domain under consideration is 
the unit square discretized by a $50\times50$
grid and with isotropic permeability field as prescribed by the
generated realizations.
To simplify the physical equations, we assume that both water and oil have
the same fluid properties (single-phase flow). Under this assumption, the
system of equations to be propagated is
\begin{align}
  -\nabla\cdot(a\nabla p) &= q \label{eq:pressure} \\
  \varphi\frac{\partial s}{\partial t} + \nabla\cdot(sv) &= q_w \label{eq:saturation}
\end{align}
where $p$ denotes the fluid pressure, $q=q_w+q_o$ denotes (total)
fluid sources, $q_w$ and $q_o$ are the water and oil sources, respectively,
$a$ denotes the permeability, $s$ denotes the saturation of water, $\varphi$
denotes the porosity, and $v$ denotes the Darcy velocity. Note that
$\mathbf a = \exp(\mathbf y)$ where $\mathbf a$ denotes the discretized
version (recall that the values of $\mathbf y$ are the
\emph{log}-permeability values).

The pressure~\cref{eq:pressure} and the
saturation~\cref{eq:saturation} are coupled through the Darcy velocity
$v=-a\nabla p$. In practice, the pressure is first solved to obtain
$v$, then the saturation is solved iteratively in time.

We consider two flow problems with different injection and
production conditions:
{\setlist[description]{font=\normalfont\itshape\space}
\begin{description}
\item[Quarter-five spot problem:] In this problem, injection and
  production points are located at $(0,0)$ and $(1,1)$ of the unit
  square, respectively. No-flow boundary conditions are imposed on all
  sides of the square. We assume unit injection/production rates,
  i.e. $q(0,0)=1$ and $q(1,1)=-1$.
\item[Uniform flow problem:] In this problem, uniformly distributed
  inflow and outflow conditions are imposed on the left and right
  sides of the unit square, respectively. No-flow boundary conditions
  are imposed on the remaining top and bottom sides. A total
  inflow/outflow rate of $1$ is assumed. For the unit square, this
  means $v\cdot\hat{n}=-1$ and $v\cdot\hat{n}=1$ on the left and right
  sides, respectively, where $\hat{n}$ denotes the outward-pointing
  unit normal to the boundary.
\end{description}
}

In both conditions, a pressure value of $0$ is imposed at the center
of the square to close the problem, the reservoir only contains oil at
the beginning (i.e. $s(\mathbf x,t=0) = 0$), and the porosity
is homogeneous and constant, with value $\varphi=0.2$. The simulated
time is from $t=0$ until $t=0.4$.  In reservoir engineering, it is
common practice to express changes in terms of pore volume injected
(PVI), defined as $\text{PVI} = \int q_{\text{in}}dt/V_{\varphi}$,
where $q_{\text{in}}$ is the water injection rate and $V_{\varphi}$ is
the pore volume of the reservoir. For a constant injection rate, this
is simply a scaling factor on the time.

For each flow problem and for each permeability pattern,
\Cref{eq:pressure,eq:saturation} were propagated for each of the three sets of
$5000$ realizations: one set corresponding to data ($5000$
randomly selected from the $40,000$ original realizations), and
the other two sets corresponding to realizations generated by PCA and
GAN.


\begin{figure}
  \centering{
  \begin{subfigure}{0.95\linewidth} \centering
    \includegraphics[width=\linewidth]{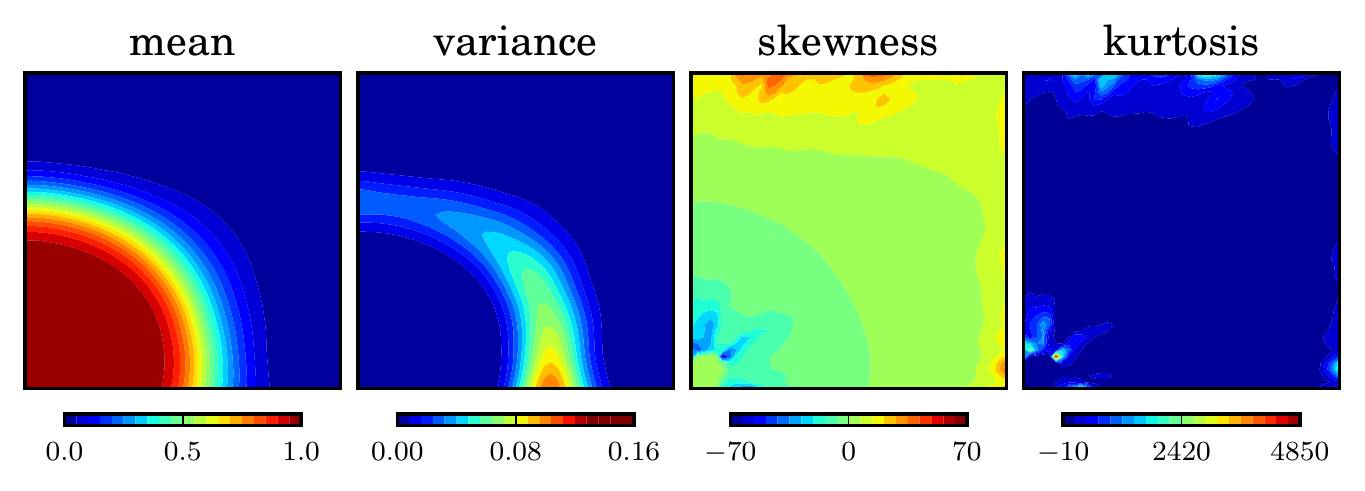}
    \caption{Statistics based on original realizations}
  \end{subfigure}
  \begin{subfigure}{0.95\linewidth} \centering
    \includegraphics[width=\linewidth]{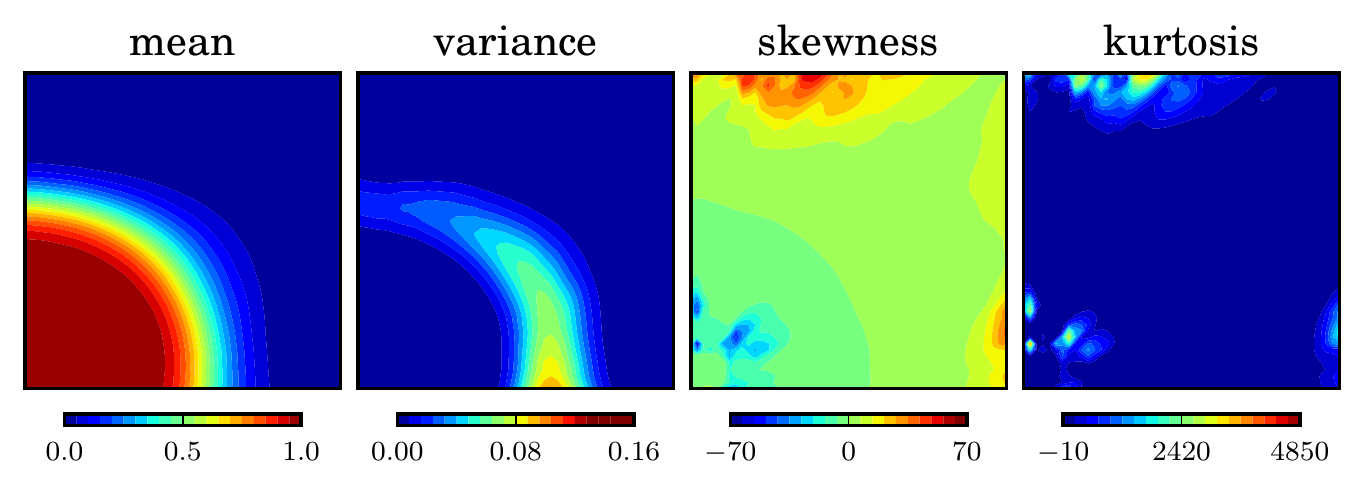}
    \caption{Statistics based on GAN realizations}
  \end{subfigure}
  \begin{subfigure}{0.95\linewidth} \centering
    \includegraphics[width=\linewidth]{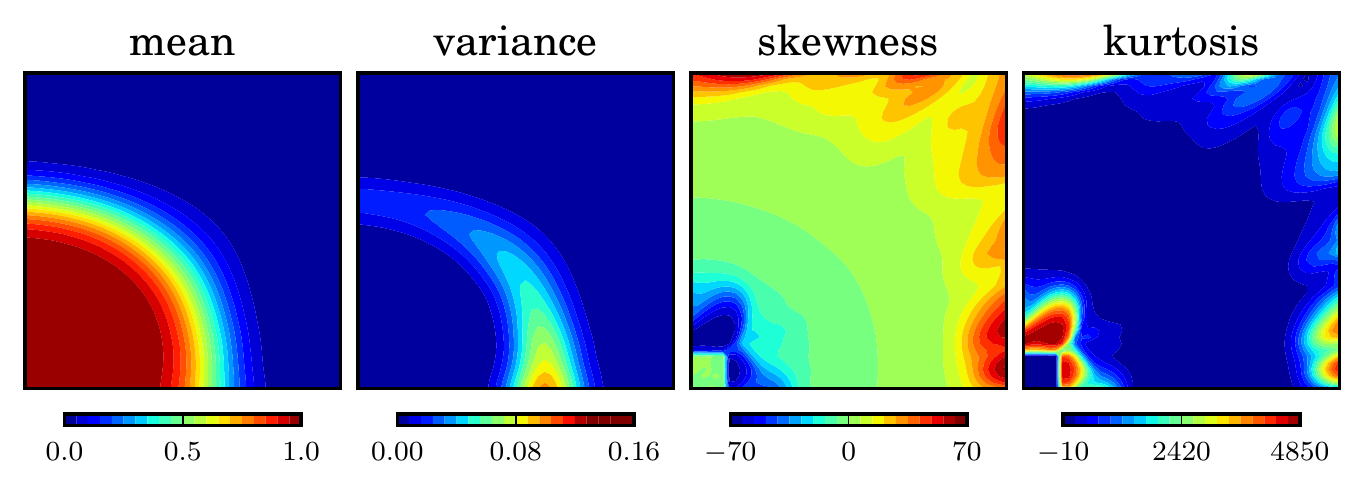}
    \caption{Statistics based on PCA realizations}
  \end{subfigure}
}
  \caption{\textbf{Quarter-five spot problem:} Statistics of the saturation map at $t=0.25\,\text{PVI}$ for
    the \emph{semi-straight} pattern.}
  \label{fig:quarter_five_channel_saturation}
\end{figure}

\begin{figure}
  \centering{
  \begin{subfigure}{0.95\linewidth} \centering
    \includegraphics[width=\linewidth]{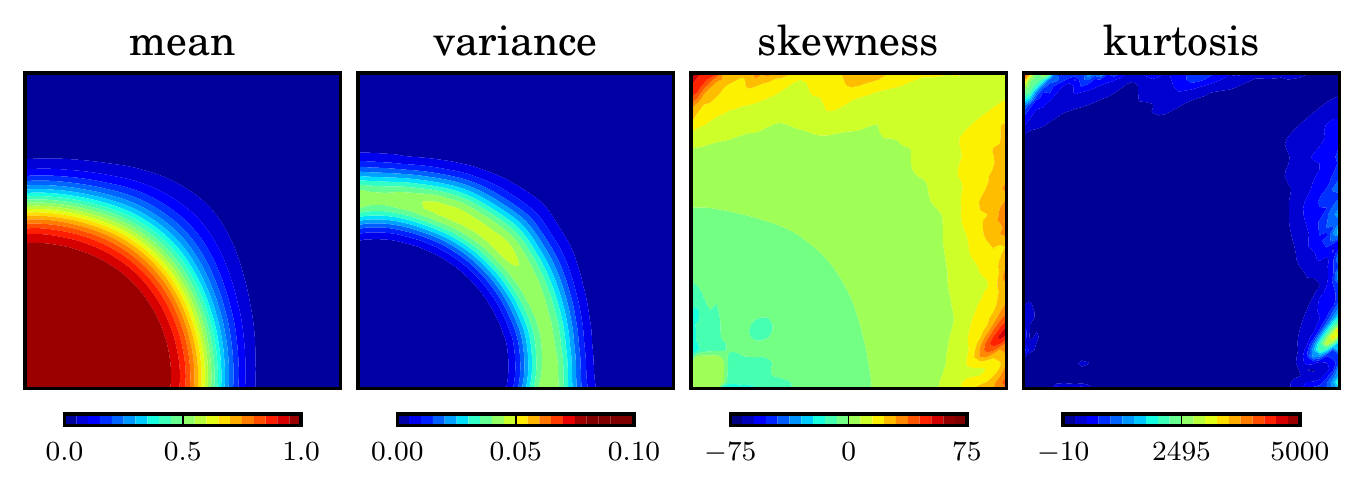}
    \caption{Statistics based on original realizations}
  \end{subfigure}
  \begin{subfigure}{0.95\linewidth} \centering
    \includegraphics[width=\linewidth]{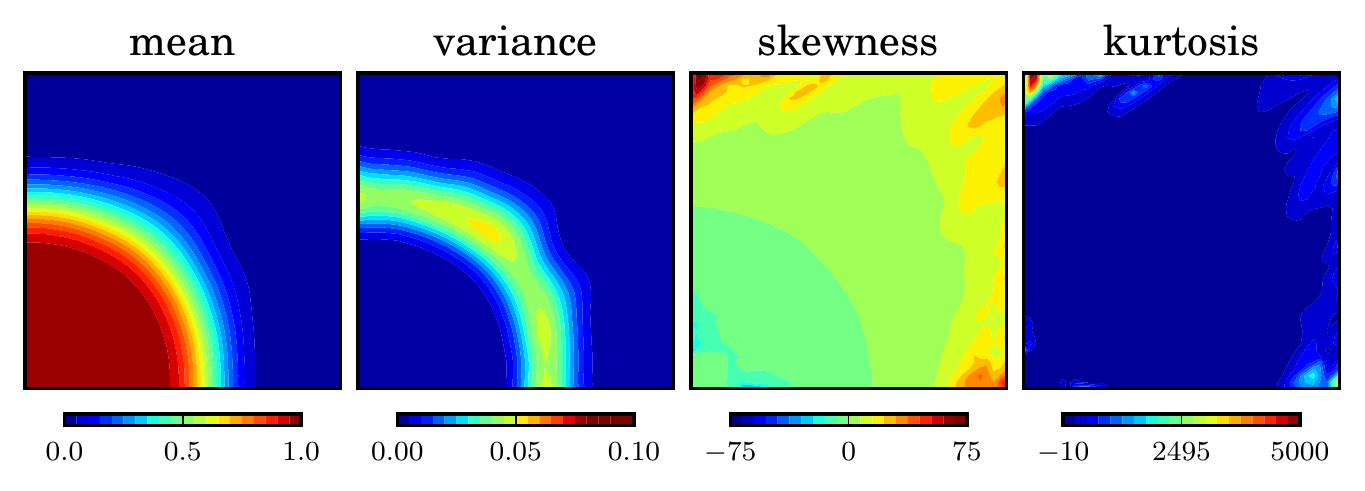}
    \caption{Statistics based on GAN realizations}
  \end{subfigure}
  \begin{subfigure}{0.95\linewidth} \centering
    \includegraphics[width=\linewidth]{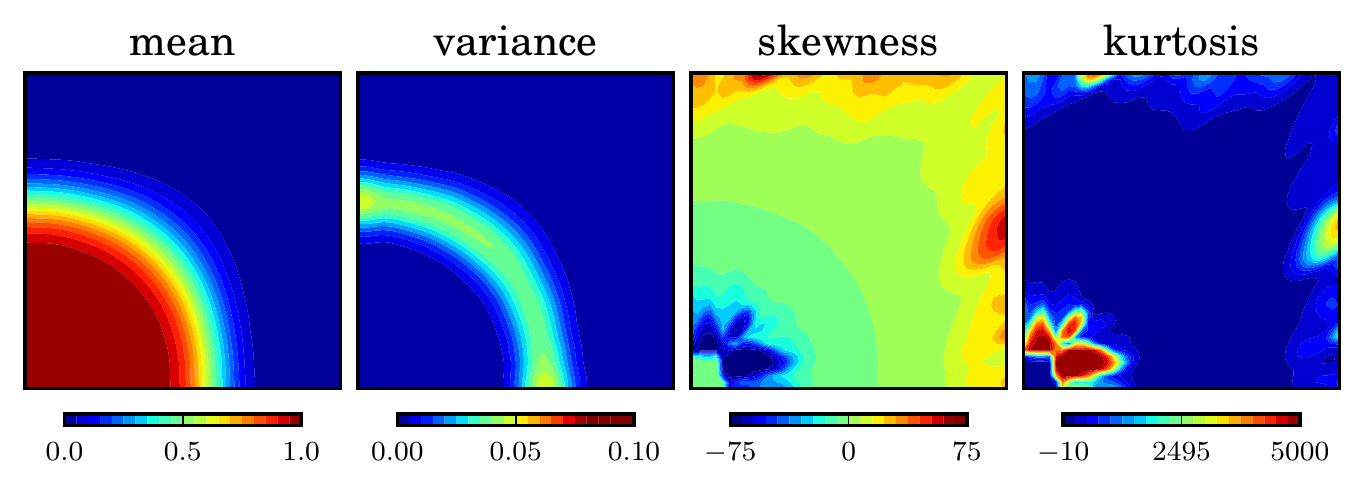}
    \caption{Statistics based on PCA realizations}
  \end{subfigure}
 }
  \caption{\textbf{Quarter-five spot problem:} Statistics of the saturation map at $t=0.25\,\text{PVI}$ for
    the \emph{meandering} pattern.}
  \label{fig:quarter_five_meandering_saturation}
\end{figure}

\begin{figure}
  \centering{
  \begin{subfigure}{0.95\linewidth} \centering
    \includegraphics[width=\linewidth]{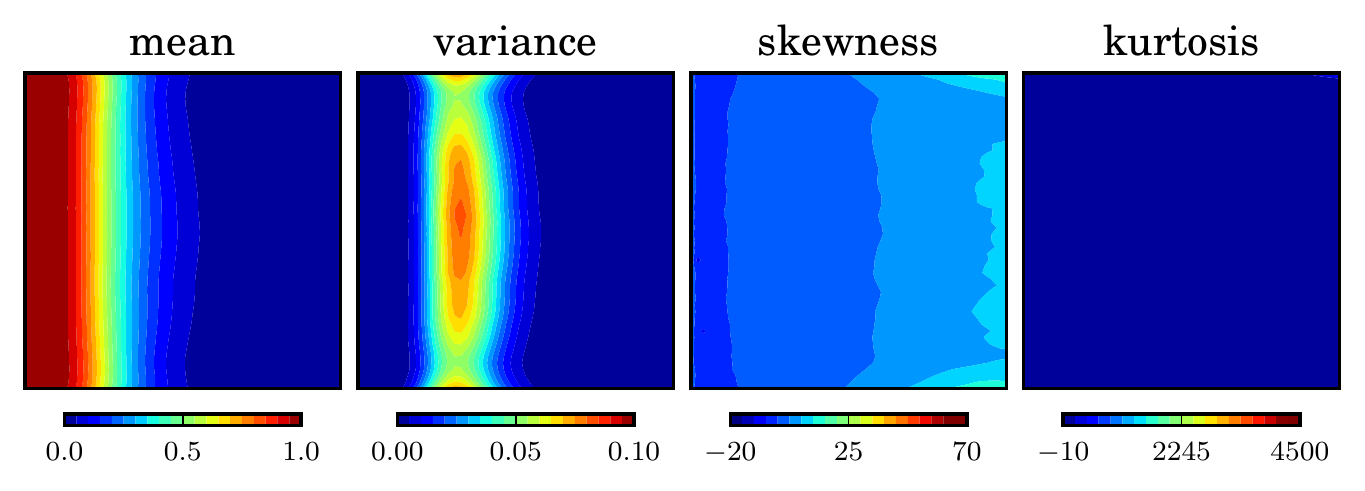}
    \caption{Statistics based on original realizations}
  \end{subfigure}
  \begin{subfigure}{0.95\linewidth} \centering
    \includegraphics[width=\linewidth]{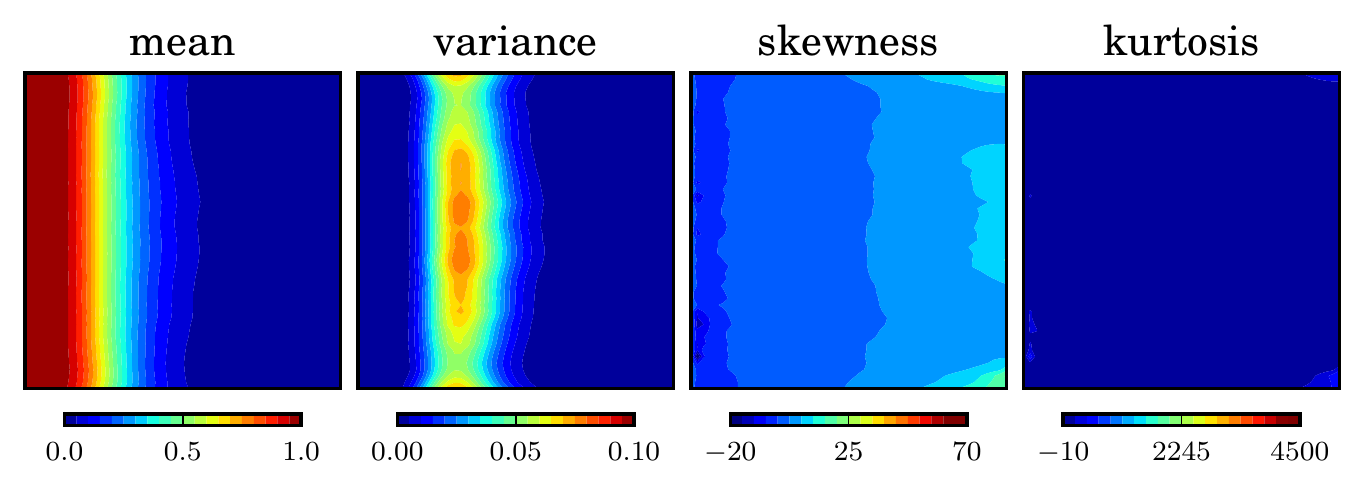}
    \caption{Statistics based on GAN realizations}
  \end{subfigure}
  \begin{subfigure}{0.95\linewidth} \centering
    \includegraphics[width=\linewidth]{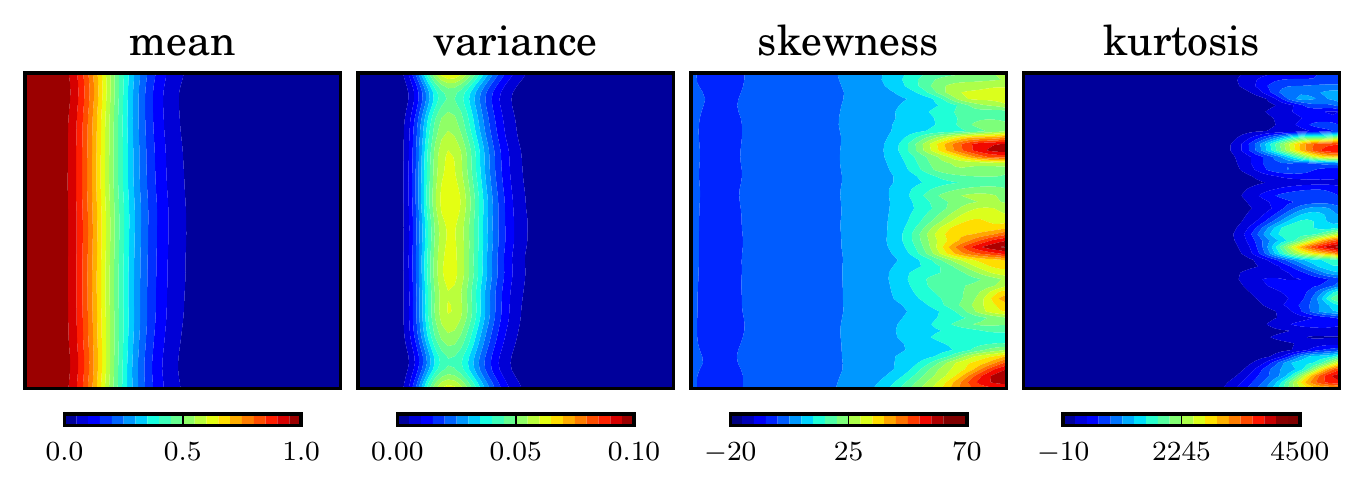}
    \caption{Statistics based on PCA realizations}
  \end{subfigure}
  }
  \caption{\textbf{Uniform flow problem:} Statistics of the saturation map at $t=0.25\,\text{PVI}$ for
    the \emph{semi-straight} pattern.}
  \label{fig:uniform_flow_channel_saturation}
\end{figure}

\begin{figure}
  \centering{
  \begin{subfigure}{0.95\linewidth} \centering
    \includegraphics[width=\linewidth]{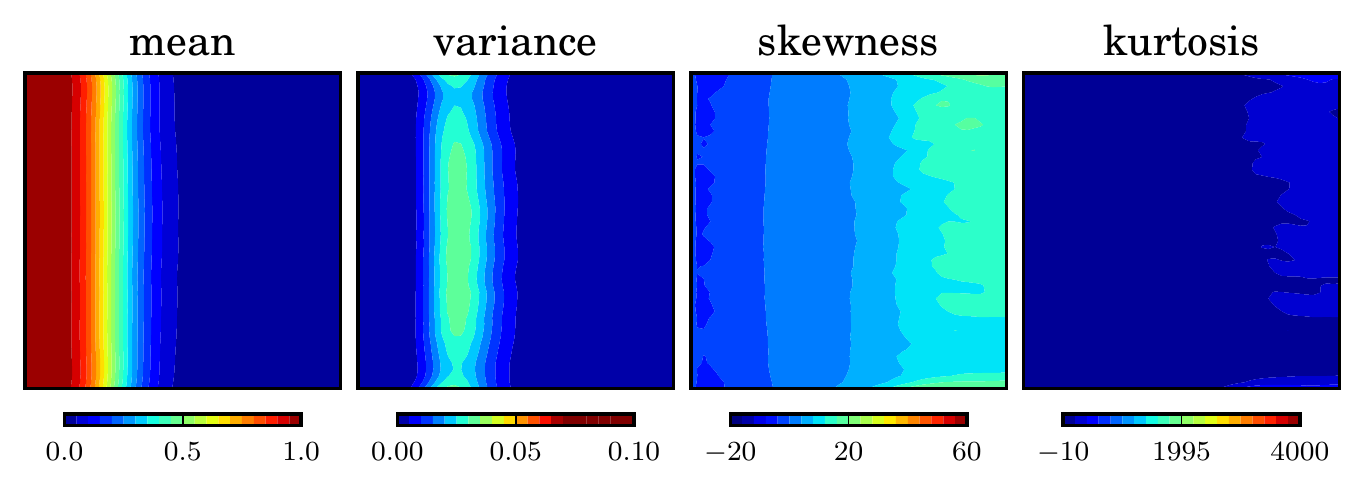}
    \caption{Statistics based on original realizations}
  \end{subfigure}
  \begin{subfigure}{0.95\linewidth} \centering
    \includegraphics[width=\linewidth]{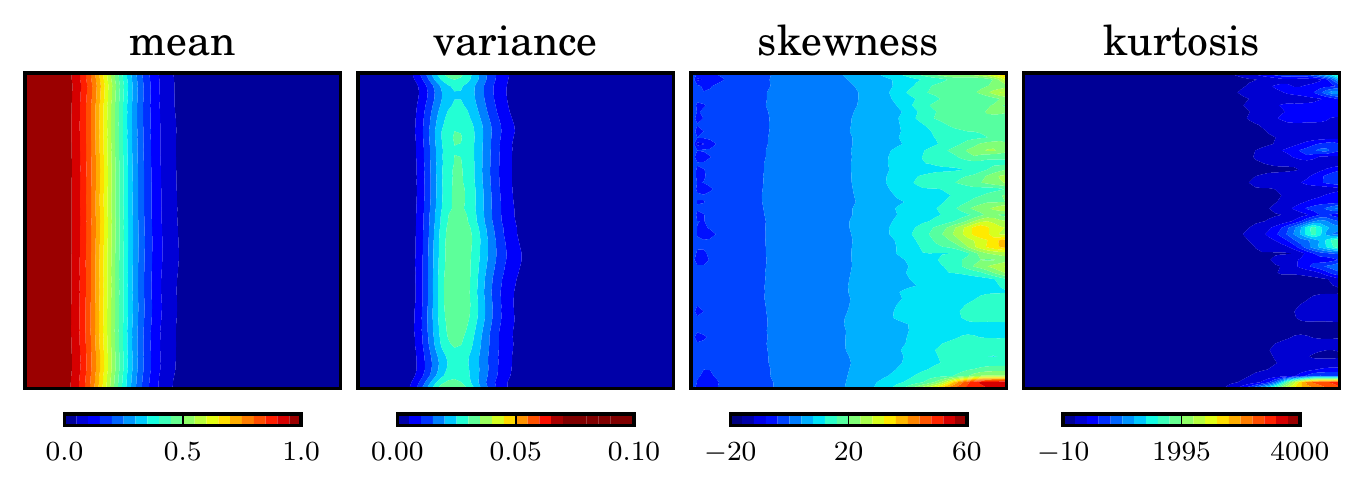}
    \caption{Statistics based on GAN realizations}
  \end{subfigure}
  \begin{subfigure}{0.95\linewidth} \centering
    \includegraphics[width=\linewidth]{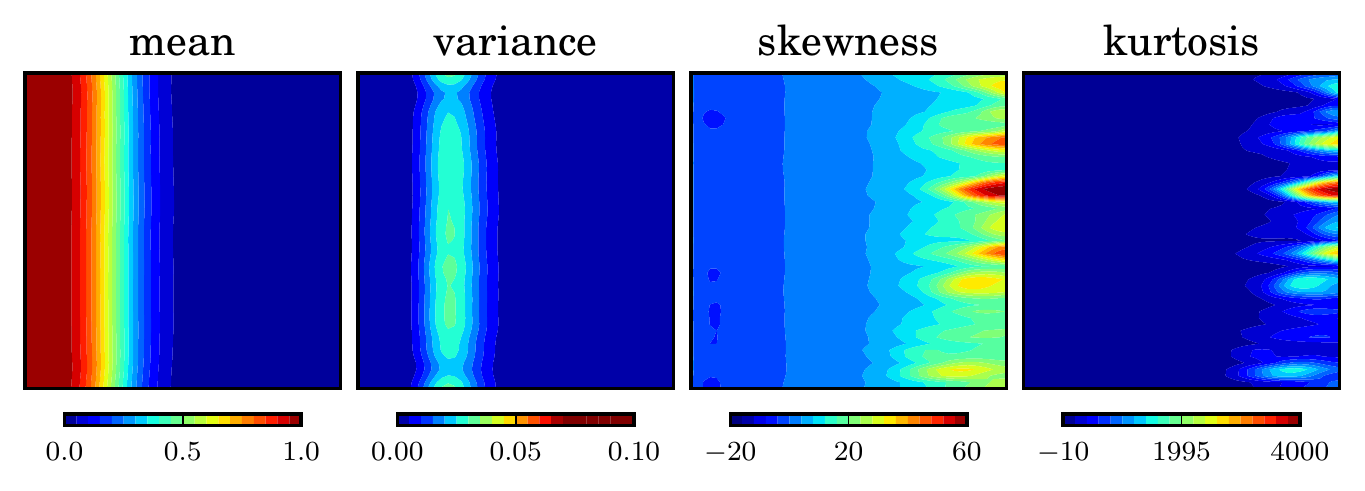}
    \caption{Statistics based on PCA realizations}
  \end{subfigure}
  }
  \caption{\textbf{Uniform flow problem:} Statistics of the saturation map at $t=0.25\,\text{PVI}$ for
    the \emph{meandering} pattern.}
  \label{fig:uniform_flow_meandering_saturation}
\end{figure}

\begin{figure}
 \centering{
  \begin{subfigure}{0.95\linewidth}
  \includegraphics[width=\linewidth]{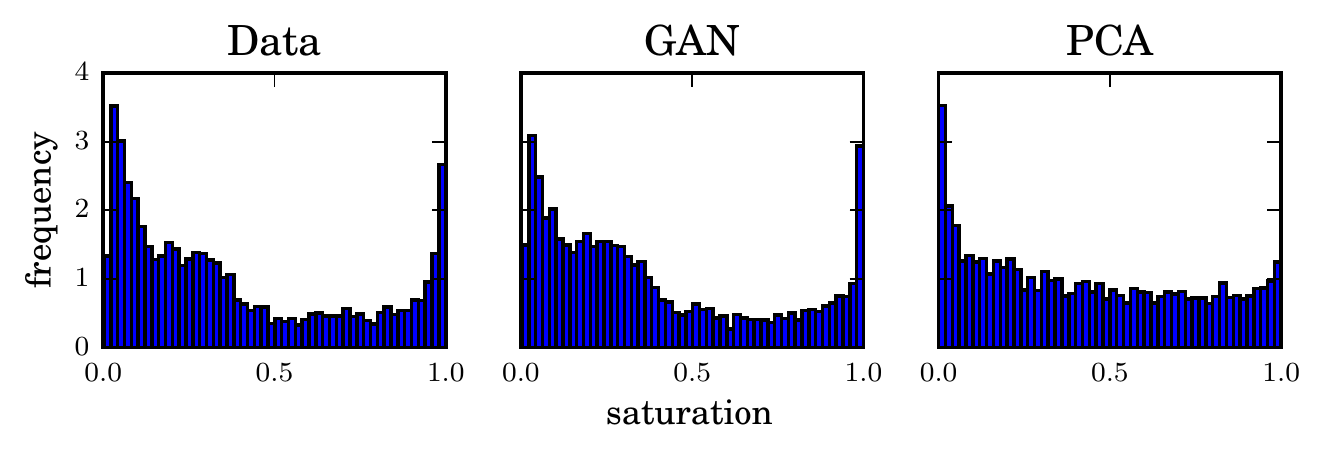}
  \caption{\emph{Semi-straight} pattern}
  \end{subfigure}
  \begin{subfigure}{0.95\linewidth}
  \includegraphics[width=\linewidth]{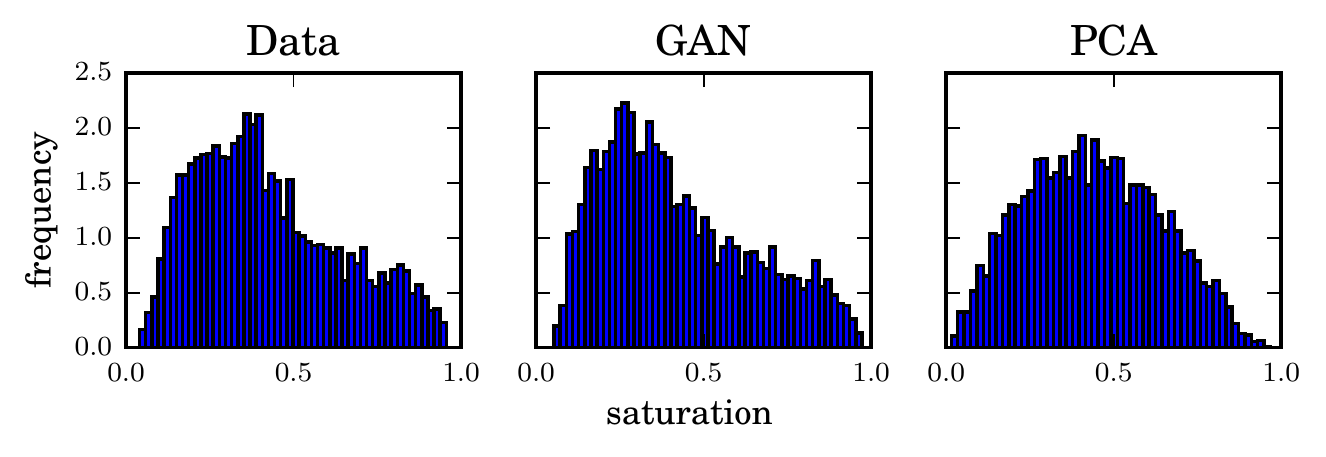}
  \caption{\emph{Meandering} pattern}
  \end{subfigure}
  }
  \caption{\textbf{Quarter-five spot problem:} Saturation histogram ($t=0.25\,\text{PVI}$) at the point of maximum variance.}
  \label{fig:quarter_five_saturation_hist}
\end{figure}

\begin{figure}
  \centering{
  \begin{subfigure}{0.95\linewidth}
  \includegraphics[width=\linewidth]{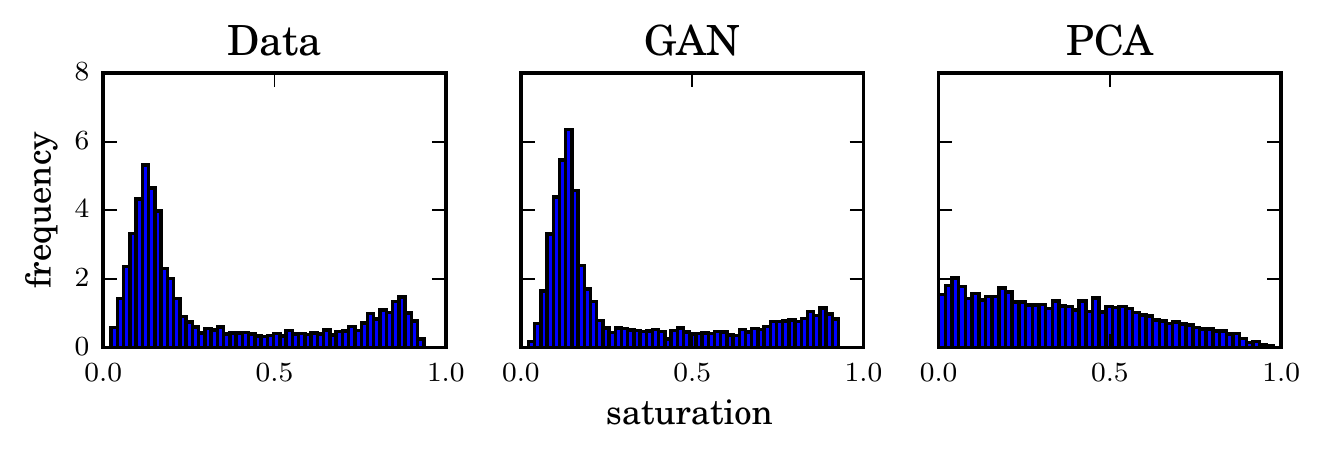}
  \caption{\emph{Semi-straight} pattern}
  \end{subfigure}
  \begin{subfigure}{0.95\linewidth}
  \includegraphics[width=\linewidth]{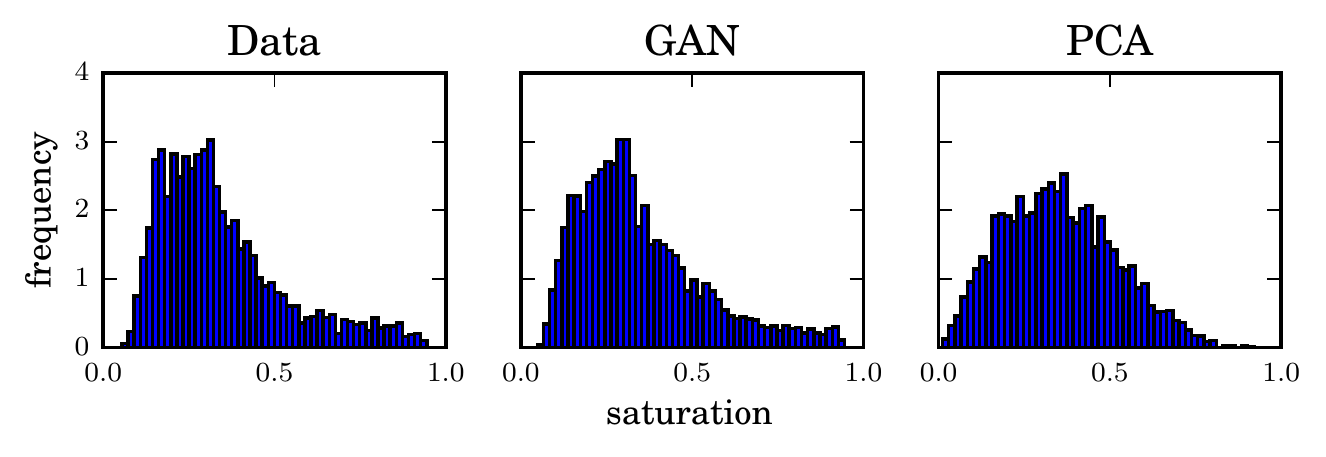}
  \caption{\emph{Meandering} pattern}
  \end{subfigure}
  }
  \caption{\textbf{Uniform flow problem:} Saturation histogram ($t=0.25\,\text{PVI}$) at the point of maximum variance.}
  \label{fig:uniform_flow_saturation_hist}
\end{figure}

After propagation, we computed the first four statistical moments
(mean, variance, skewness and kurtosis) of the saturation at time
$t=0.25\,\text{PVI}$, based on the $5000$ runs. The results are shown
in
\Cref{fig:quarter_five_channel_saturation,fig:quarter_five_meandering_saturation,fig:uniform_flow_channel_saturation,fig:uniform_flow_meandering_saturation}
for each test case.  In the mean and variance maps, both PCA and GAN
are very close to the true maps. The good performance of PCA is
expected since this method aims to preserve the first two statistical
moments. For the higher order moments (skewness and kurtosis),
estimations by GAN are overall better than PCA, as seen from the map
plots. These observations are general in all the test cases.
To further compare the statistics of the saturation solutions, we
plot the saturation histogram ($t=0.25\,\text{PVI}$) at the point of
maximum variance, shown
in~\Cref{fig:quarter_five_saturation_hist,fig:uniform_flow_saturation_hist}.
In accordance to the map plot results, we see that PCA fails to
replicate the higher order characteristics of the saturation
distribution. On the other hand, the distributions estimated from GAN are
surprisingly close to the true distributions, suggesting that it has
generated samples that preserve the original flow properties.


\begin{figure}
  \centering{
  \begin{subfigure}{0.95\linewidth}
    \begin{subfigure}{.49\linewidth}
        \includegraphics[width=\linewidth]{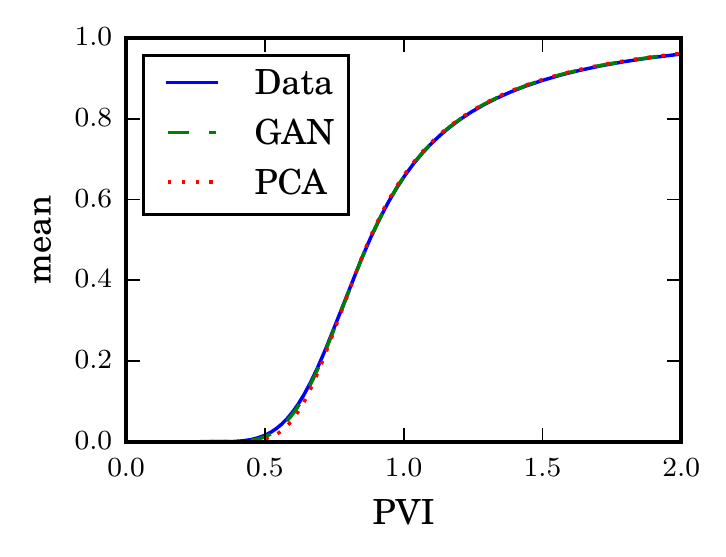} 
    \end{subfigure}
    \begin{subfigure}{.49\linewidth}
        \includegraphics[width=\linewidth]{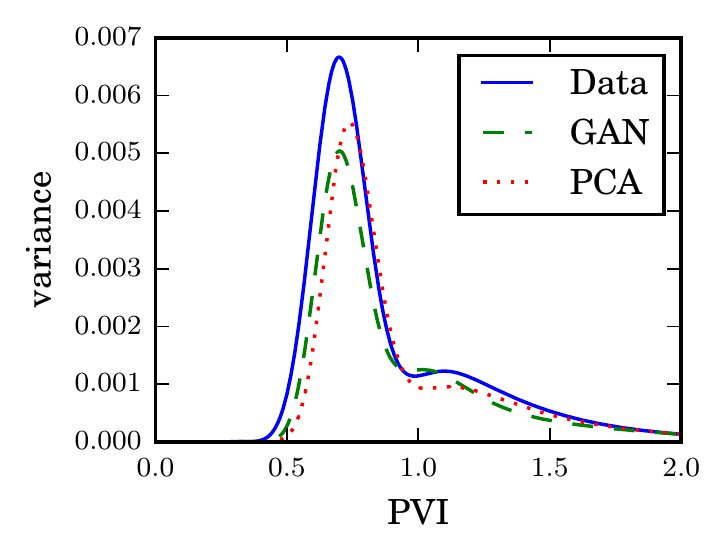} 
    \end{subfigure}
    \caption{\emph{Semi-straight} pattern}
  \end{subfigure}
  \begin{subfigure}{0.95\linewidth}
    \begin{subfigure}{.49\linewidth}
    \includegraphics[width=\linewidth]{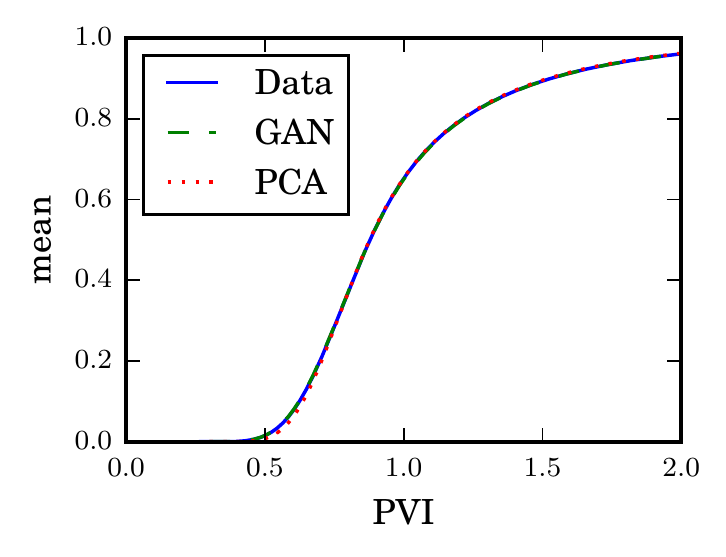} 
    \end{subfigure}
    \begin{subfigure}{.49\linewidth}
    \includegraphics[width=\linewidth]{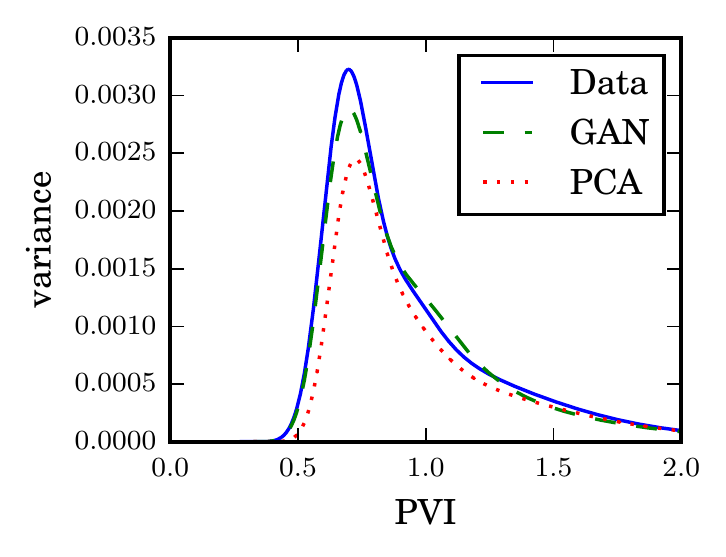} 
    \end{subfigure}
    \caption{\emph{Meandering} pattern}
  \end{subfigure}
  }
    \caption{\textbf{Quarter-five spot problem:} Mean and variance of the water cut curves.}
  \label{fig:quarter_five_wc}
\end{figure}

\begin{figure}
  \centering{
  \begin{subfigure}{0.95\linewidth}
    \begin{subfigure}{.49\linewidth}
        \includegraphics[width=\linewidth]{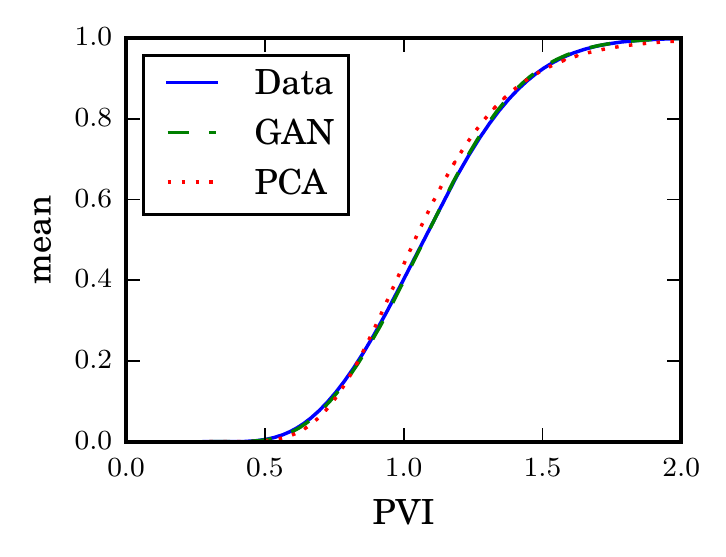} 
    \end{subfigure}
    \begin{subfigure}{.49\linewidth}
        \includegraphics[width=\linewidth]{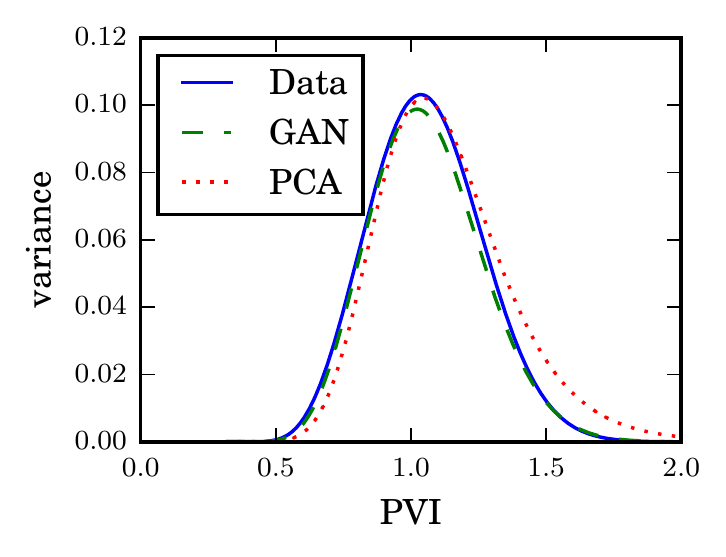} 
    \end{subfigure}
    \caption{\emph{Semi-straight} pattern}
  \end{subfigure}
  \begin{subfigure}{0.95\linewidth}
    \begin{subfigure}{.49\linewidth}
    \includegraphics[width=\linewidth]{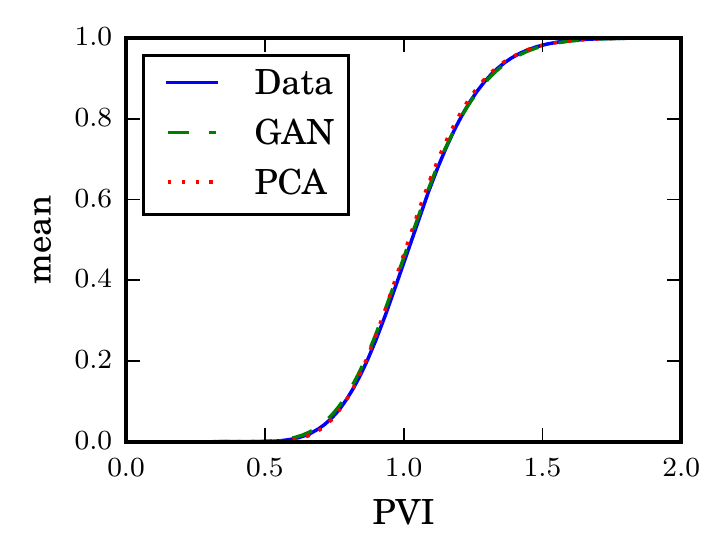} 
    \end{subfigure}
    \begin{subfigure}{.49\linewidth}
    \includegraphics[width=\linewidth]{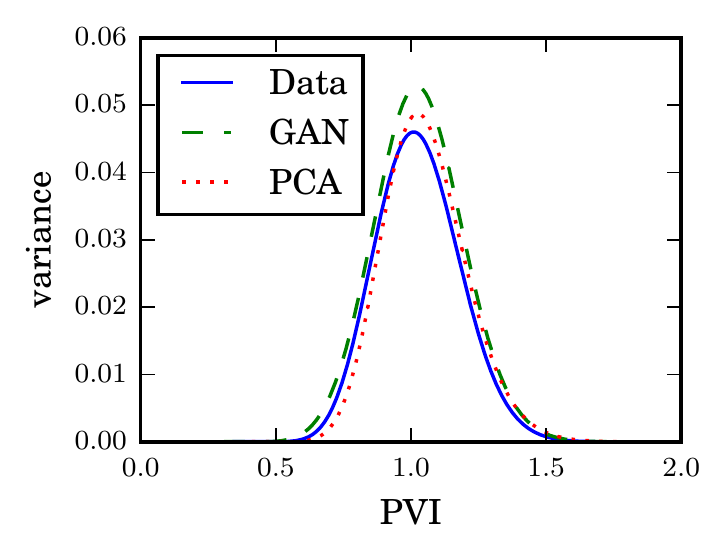} 
    \end{subfigure}
    \caption{\emph{Meandering} pattern}
  \end{subfigure}
  }
    \caption{\textbf{Uniform flow problem:} Mean and variance of the water-cut curves.}
  \label{fig:uniform_flow_wc}
\end{figure}

\begin{figure}
  \centering{
  \begin{subfigure}{0.95\linewidth}
  \includegraphics[width=\linewidth]{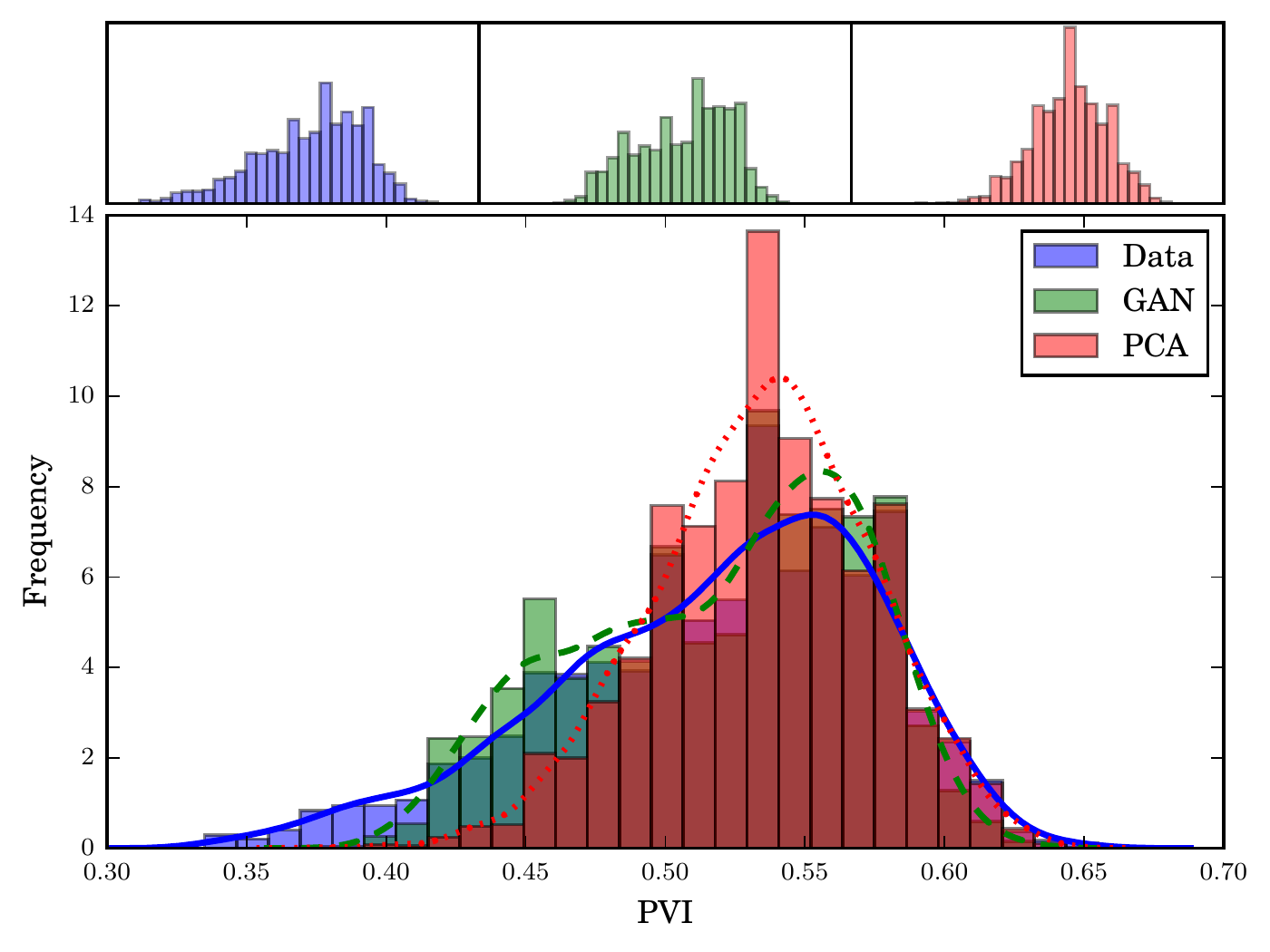}
  \caption{\emph{Semi-straight} pattern}
  \end{subfigure}
  \begin{subfigure}{0.95\linewidth}
  \includegraphics[width=\linewidth]{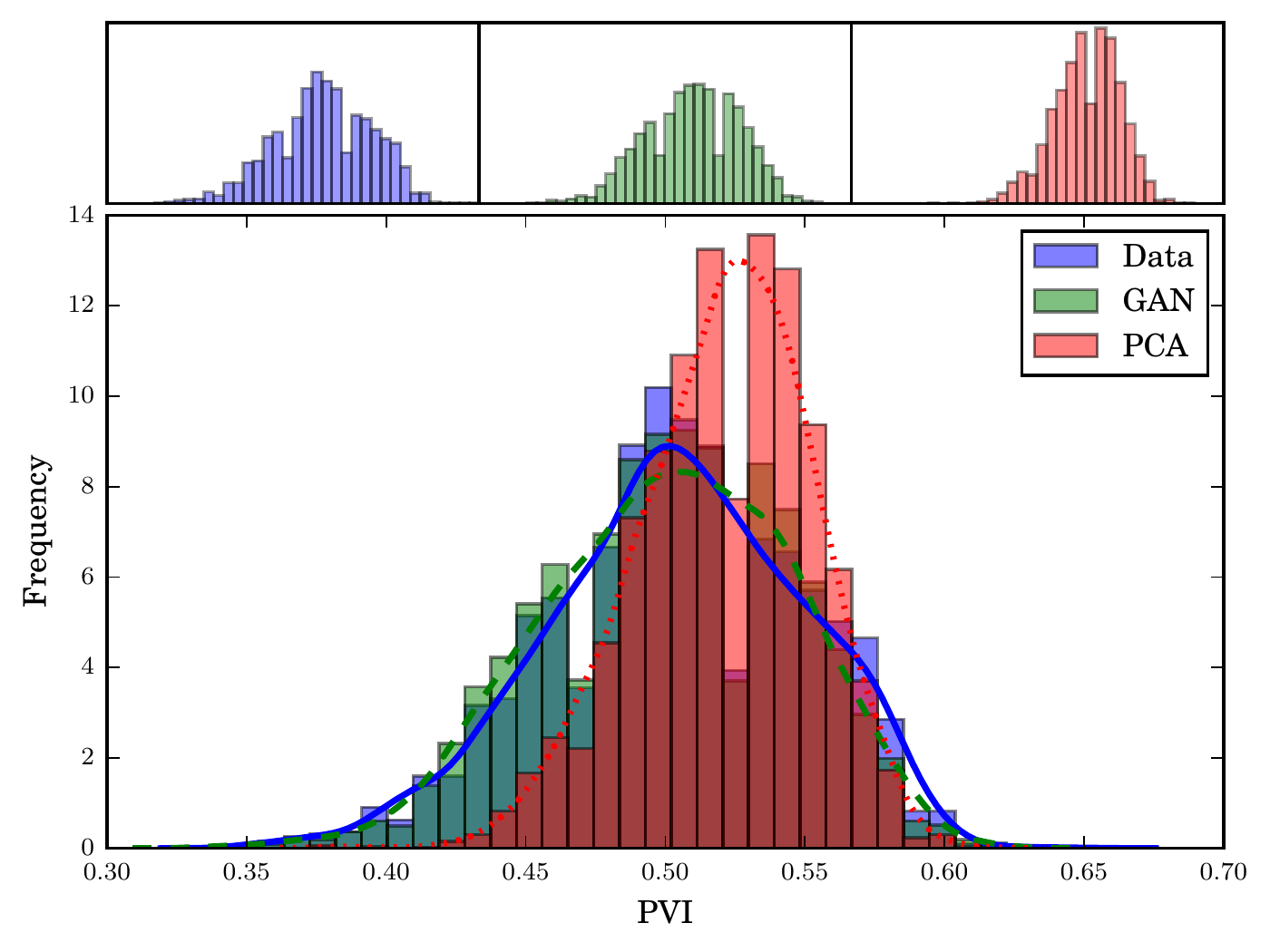}
  \caption{\emph{Meandering} pattern}
  \end{subfigure}
  }
  \caption{\textbf{Quarter-five spot problem:} Histogram and PDF of the water breakthrough times.}
  \label{fig:quarter_five_wbt}
\end{figure}

\begin{figure}
  \centering{
  \begin{subfigure}{0.95\linewidth}
  \includegraphics[width=\linewidth]{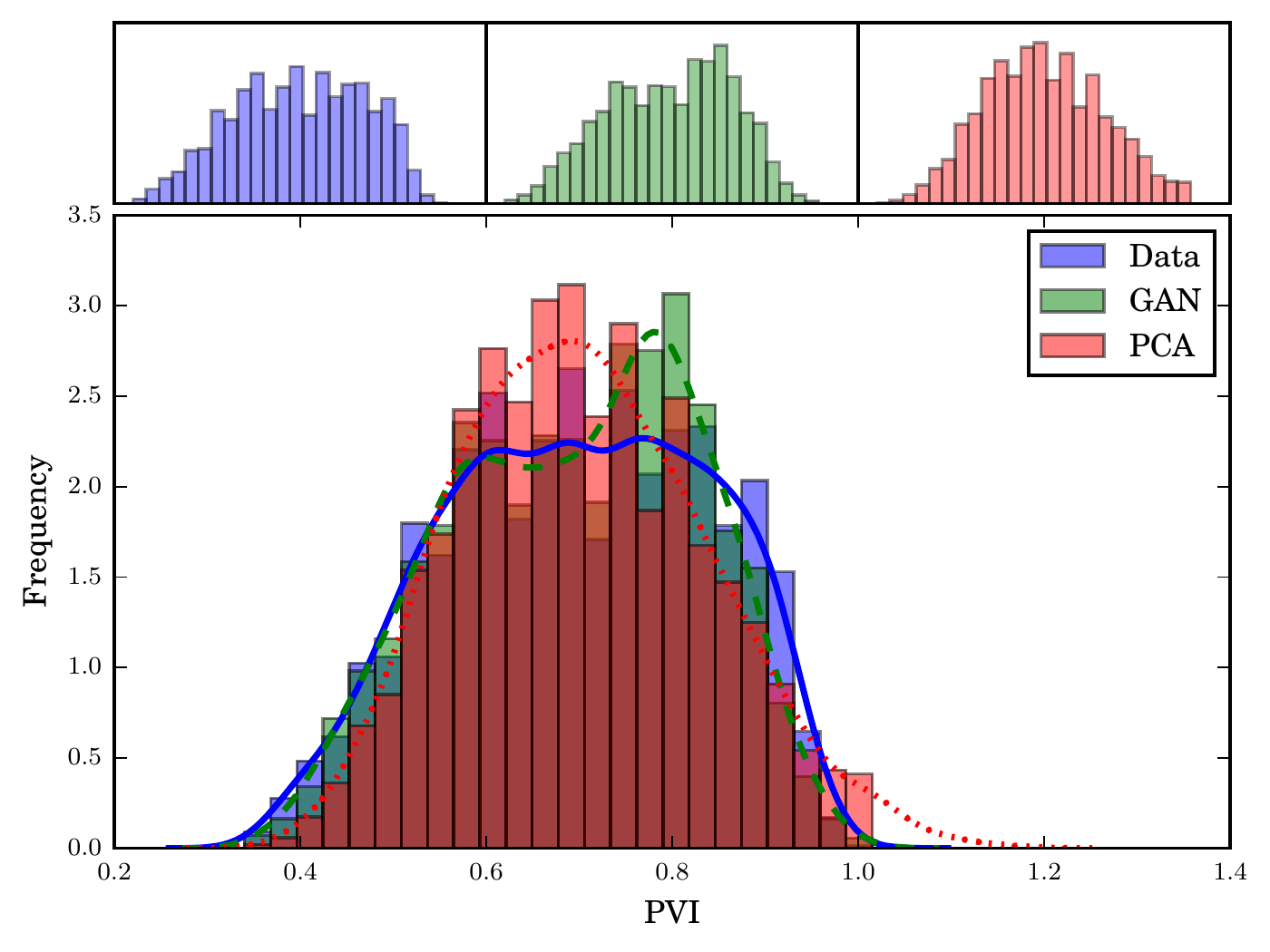}
  \caption{\emph{Semi-straight} pattern}
  \end{subfigure}
  \begin{subfigure}{0.95\linewidth}
  \includegraphics[width=\linewidth]{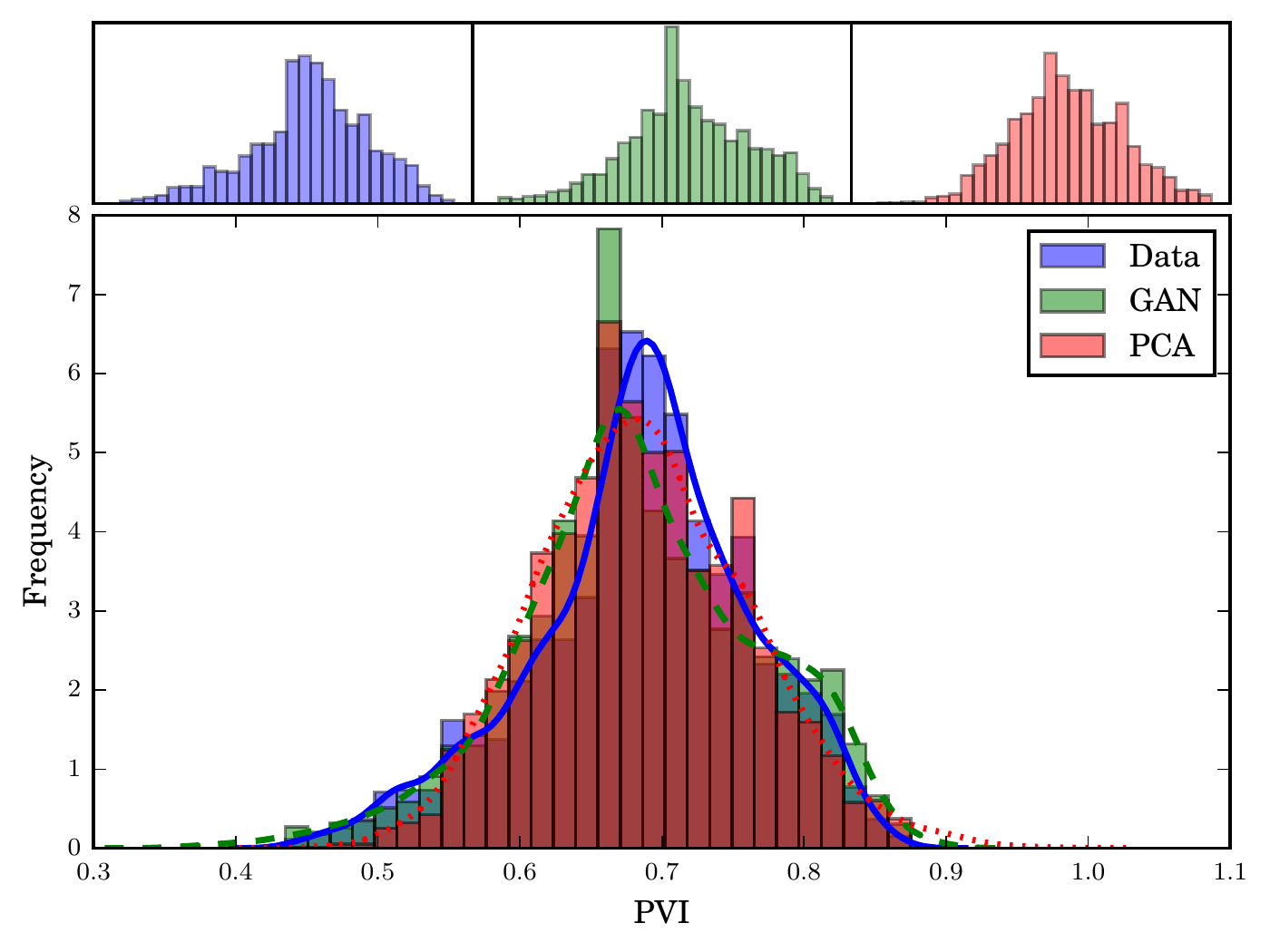}
  \caption{\emph{Meandering} pattern}
  \end{subfigure}
  }
  \caption{\textbf{Uniform flow problem:} Histogram and PDF of the water breakthrough times.}
  \label{fig:uniform_flow_wbt}
\end{figure}

Next, we look at the water-cut curves, which measure the water content
in the produced fluid over time.
We computed the mean and variance of the water-cut curves from the
$5000$ runs The results are shown in
\Cref{fig:quarter_five_wc,fig:uniform_flow_wc} for each test case.
We observe that the mean curves are hardly distinguishable in all
cases. Regarding the variance curves, two cases favored PCA
(quarter-five spot in the semi-straight pattern, and uniform flow in
the meandering pattern) while the other two cases favored GAN. We
emphasize again that PCA is expected to excel in the low-order statistics,
yet the results of GAN are comparable.

Finally, we look at the histogram of the water breakthrough time
(measured in PVI), defined as the time at which water-cut reaches 1\%
of the produced fluids. \Cref{fig:quarter_five_wbt,fig:uniform_flow_wbt} show the
histogram and estimated densities of the water
breakthrough time for the quarter-five spot problem and the uniform
flow problem, respectively. The estimated densities were obtained using
Scott's rule. In all cases, we found that the obtained densities from GAN are
in good agreement with the reference densities, clearly outperforming PCA.

\section{Conclusions}
We presented one of the first applications of \emph{generative adversarial
networks} (GAN) for the parametrization of geological models. The
results of our study show the potential of GANs as a
parametrization tool to capture complex structures in geological models.
The method generated geological realizations that were visually
plausible, reproducing the structures seen in the reference models.
More importantly, the flow statistics induced by the generated
realizations were in close agreement with the reference.
This was verified for two different permeability models for two
test cases in subsurface flow.
In particular, we performed uncertainty propagation to estimate distributions
of several quantities of interest, and found that GAN was very effective in
approximating the true densities even when the distributions
were non-trivial. In contrast, PCA was not suitable for
distributions that are far removed from the normal
distribution. Furthermore, we note that GAN showed superior
performance compared to PCA despite only employing $20$
coefficients, in contrast to PCA where $37$ and $104$ coefficients
were used (retaining 75\% of the variance) for the semi-straight and meandering patterns, respectively.

We conclude that GANs present a promising method for the
parametrization of geological models.  In our future work, we will
look into the practicality of GANs for inverse modeling where the
differentiability of the generator could be exploited to build an
efficient gradient-based inverse method.

\bibliographystyle{plainnat}
\bibliography{biblio}

\pagebreak
\appendix
\section{Additional samples} \label{sec:appendix}

\begin{figure}[!htb] \centering
  \begin{subfigure}{.95\linewidth} \centering
    \includegraphics[width=\linewidth]{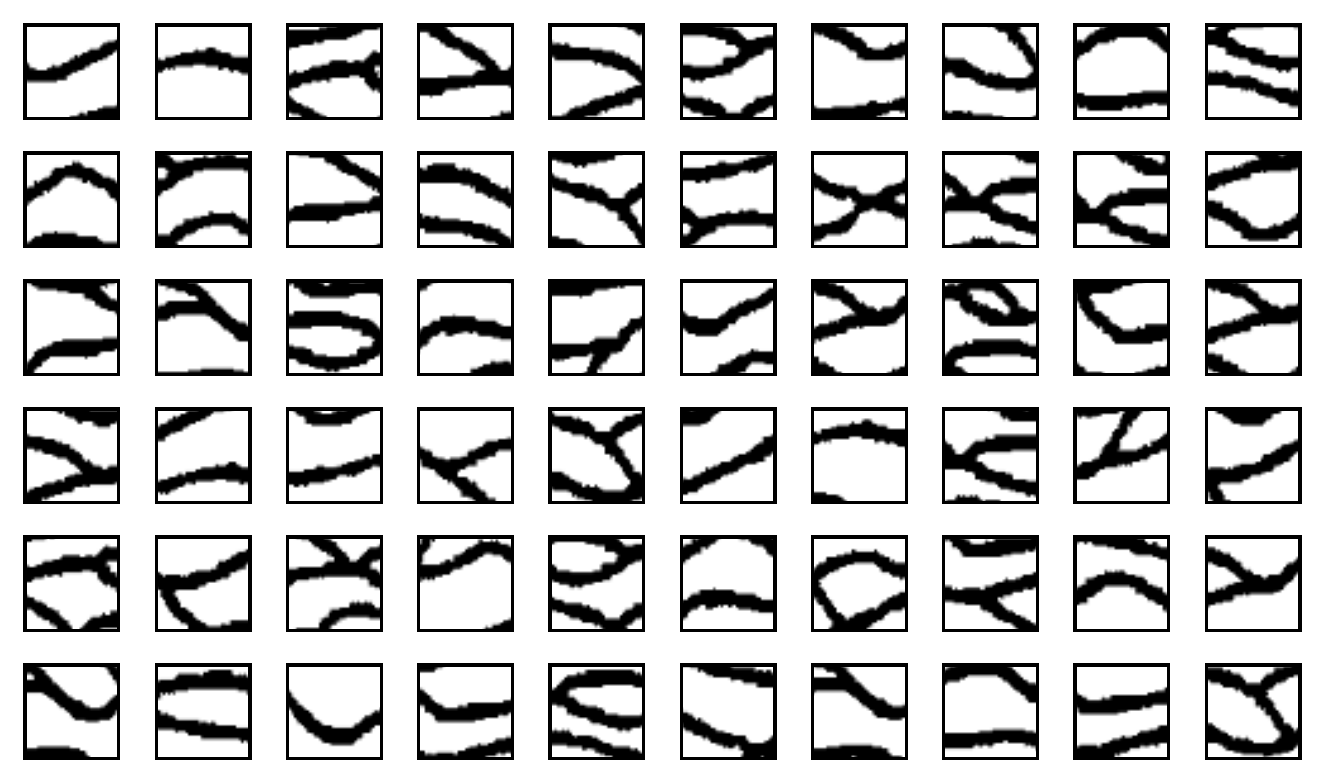}
    \caption{\emph{Semi-straight} channels}
  \end{subfigure}
  \begin{subfigure}{.95\linewidth} \centering
    \includegraphics[width=\linewidth]{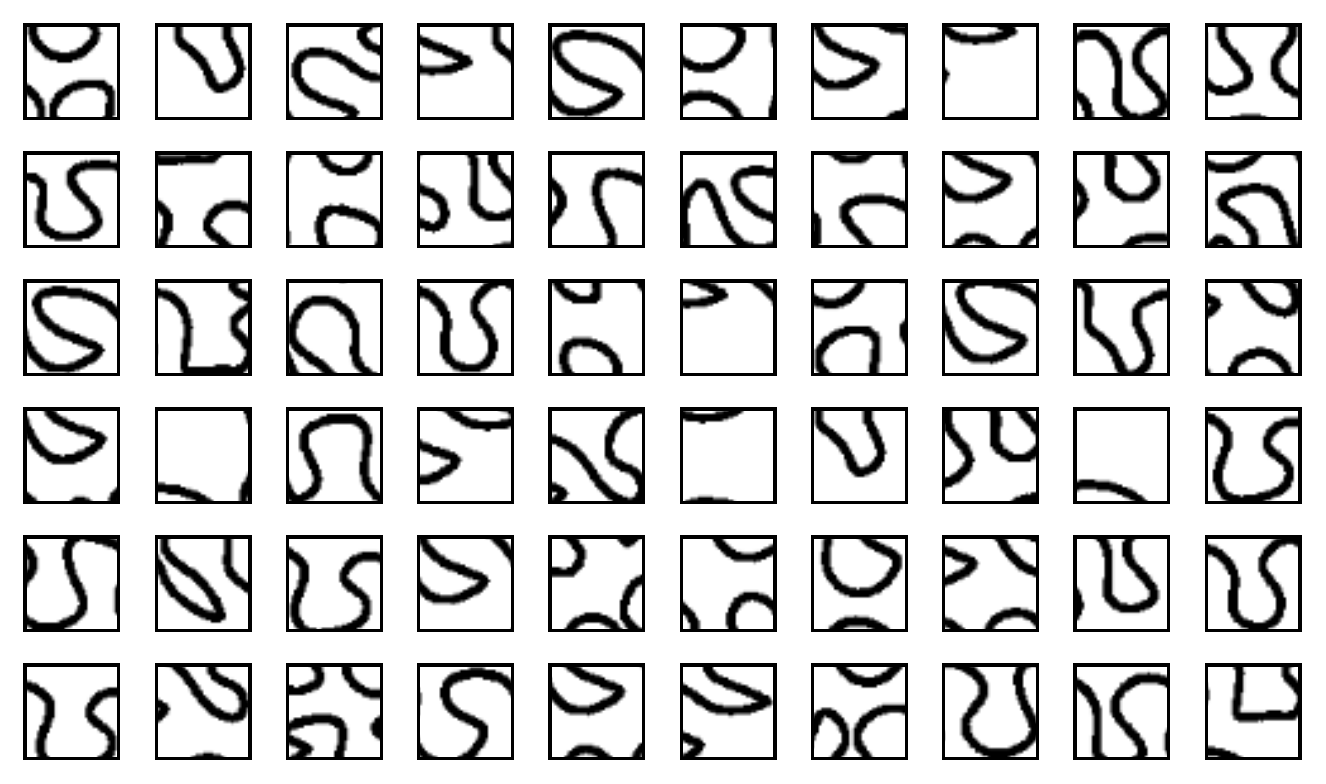}
    \caption{\emph{Meandering} channels}
  \end{subfigure}
  \caption{Original realizations}
\end{figure}

\begin{figure}[!htb] \centering
  \begin{subfigure}{.95\linewidth} \centering
    \includegraphics[width=\linewidth]{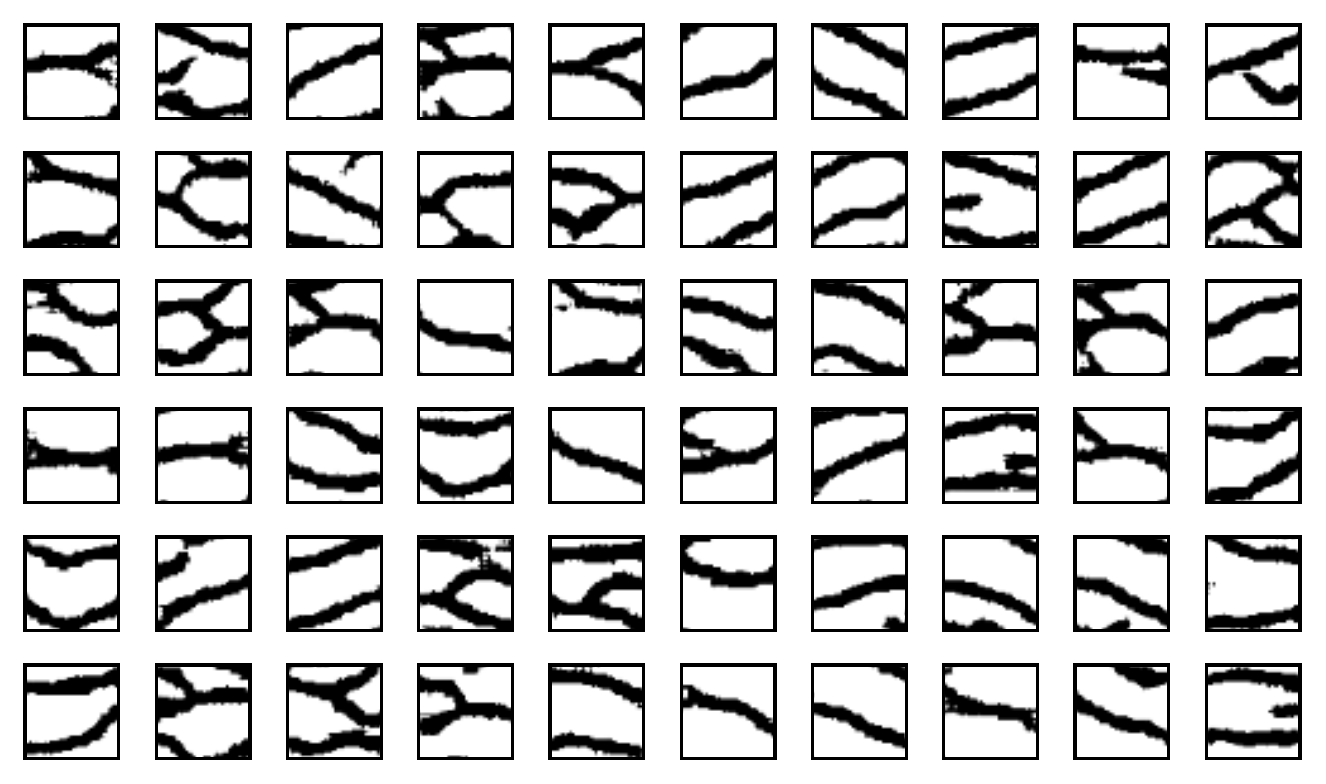}
    \caption{\emph{Semi-straight} channels}
  \end{subfigure}
  \begin{subfigure}{.95\linewidth} \centering
    \includegraphics[width=\linewidth]{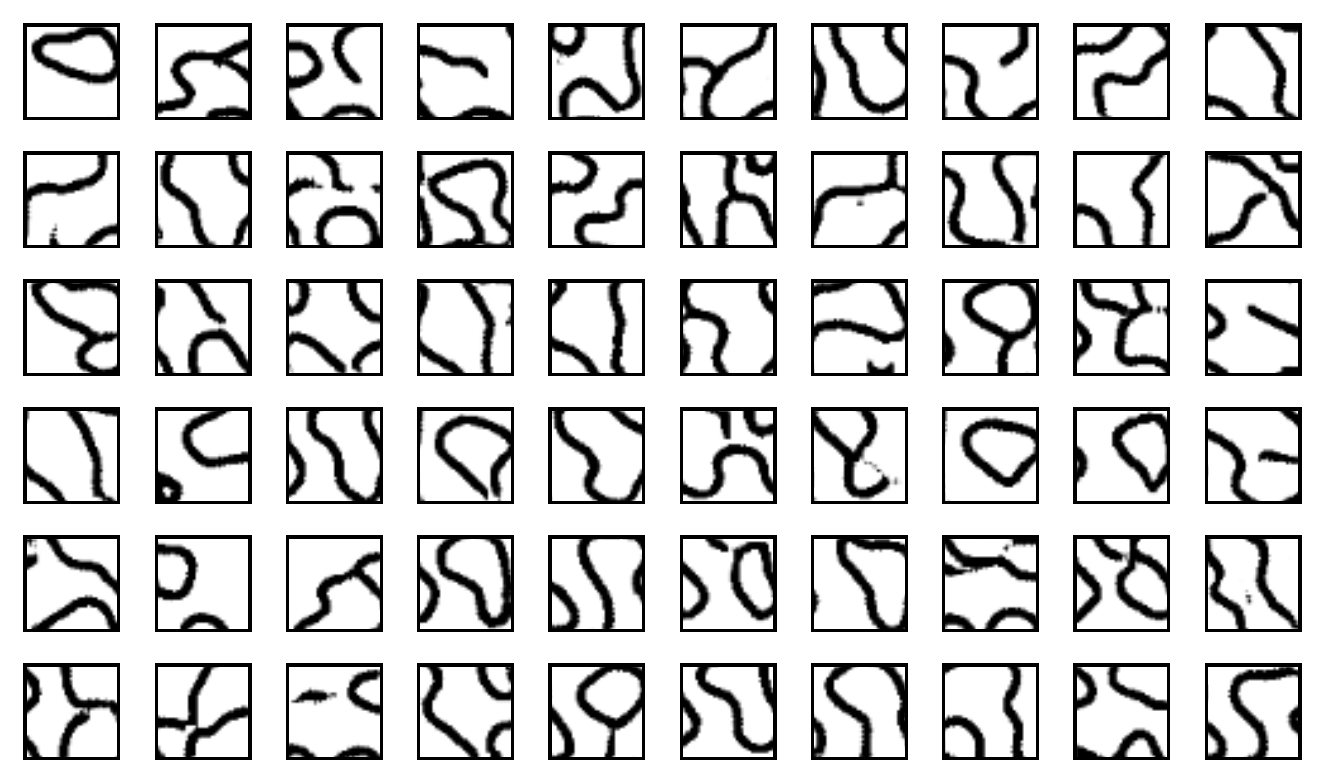}
    \caption{\emph{Meandering} channels}
  \end{subfigure}
  \caption{Realizations by GAN20}
\end{figure}

\begin{figure}[!htb] \centering
  \begin{subfigure}{.95\linewidth} \centering
    \includegraphics[width=\linewidth]{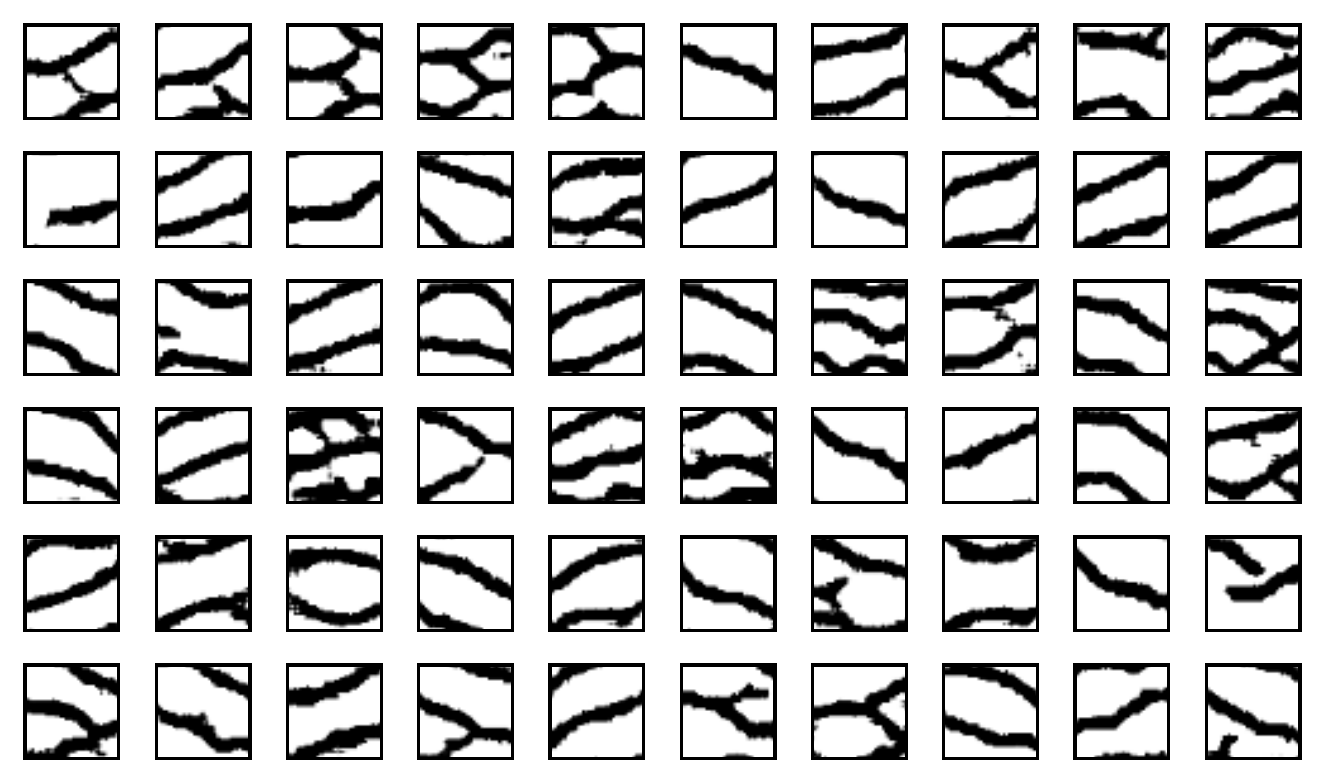}
    \caption{\emph{Semi-straight} channels}
  \end{subfigure}
  \begin{subfigure}{.95\linewidth} \centering
    \includegraphics[width=\linewidth]{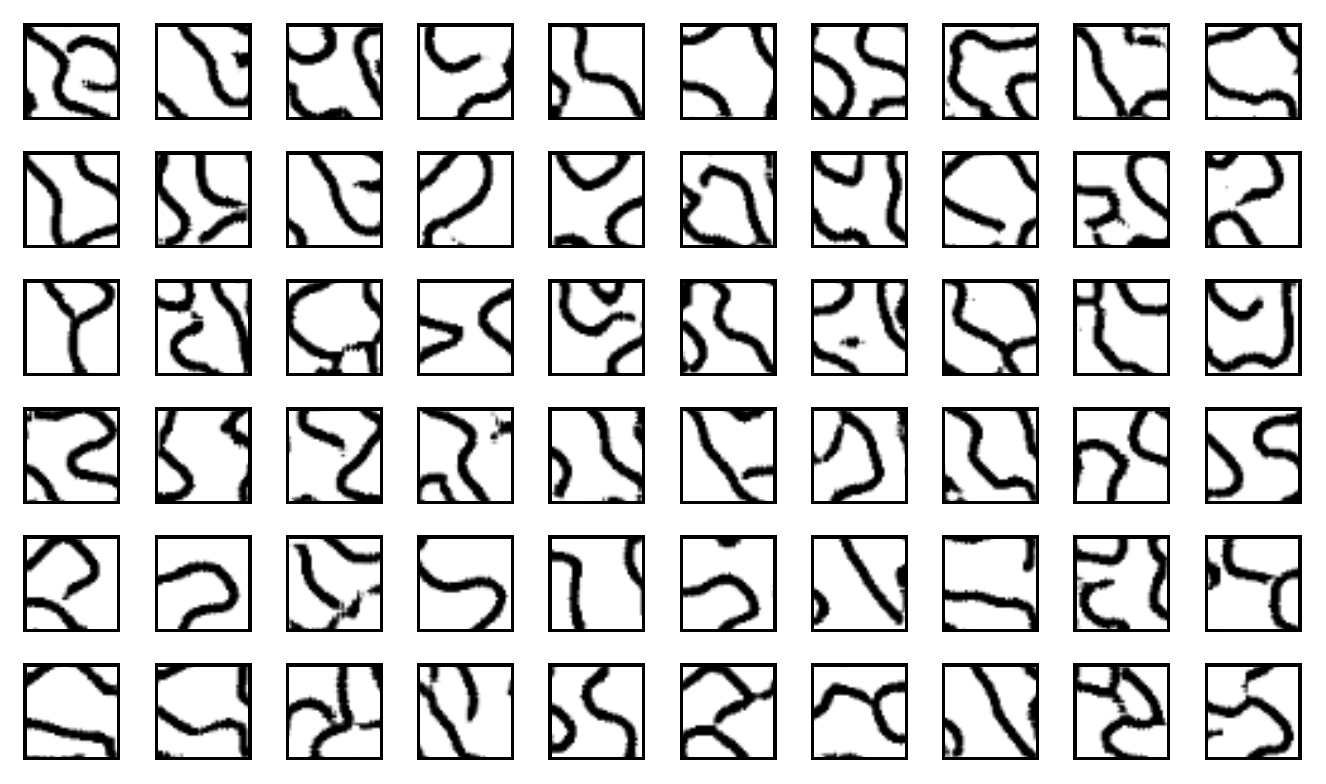}
    \caption{\emph{Meandering} channels}
  \end{subfigure}
  \caption{Realizations by GAN40}
\end{figure}

\end{document}